\documentclass[journal]{IEEEtran}

\usepackage{cite}

\ifCLASSINFOpdf
   \usepackage[pdftex]{graphicx}
\else
\fi

\usepackage[colorlinks=true,linkcolor=black,citecolor=black]{hyperref}
\usepackage{mathrsfs}
\usepackage{amssymb}
\usepackage{amsmath}
\usepackage{array}
\usepackage{multirow}

\usepackage{algorithm}  
\usepackage{algpseudocode}  
\usepackage{amsmath}  
\usepackage{color}
\newtheorem{remark}{Remark}

\begin{document}
	
\bstctlcite{BSTcontrol}	
\title{Cell Biomechanical Modeling Based on Membrane Theory with Considering Speed Effect of Microinjection}

\author{Shengzheng~Kang, Zhicheng~Song, Xiaolong~Yang, Yao~Li, 
		  Hongtao~Wu, and~Tao~Li
\thanks{This work was supported by the National Natural Science Foundation of China under Grant 51975277, the Natural Science Foundation of Jiangsu Province under Grant BK20220449, and the Startup Foundation for Introducing Talent of NUIST under Grant 2022r092. 
\textit{(Corresponding author: Tao Li, Shengzheng Kang.)}
}
\thanks{S. Kang and T. Li are with the School of Automation, Nanjing University of Information Science and Technology, 219 Ningliu Road, Pukou District, Nanjing 210044, China (e-mail: kangsz@nuist.edu.cn; litaojia@nuist.edu.cn).}
\thanks{Z. Song and H. Wu are with the College of Mechanical and Electrical Engineering, Nanjing University of Aeronautics and Astronautics, 29 Yudao Street, Qinhuai District, Nanjing 210016, China (e-mail: song\_zhicheng@nuaa.edu.cn; meehtwu@nuaa.edu.cn).}
\thanks{X. Yang is with the School of Mechanical Engineering, Nanjing University of Science and Technology, 200 Xiaolingwei Street, Xuanwu District, Nanjing 210094, China (e-mail: xiaolongyang@njust.edu.cn).}
\thanks{Y. Li is with the Industrial Center, School of Innovation and Entrepreneurship, Nanjing Institute of Technology, 1 Hongjing Avenue, Jiangning District, Nanjing 211167, China (e-mail: liyaokkx@njit.edu.cn).}
}

\maketitle

\begin{abstract}
As an effective method to deliver external materials into biological cells, microinjection has been widely applied in the biomedical field. However, the cognition of cell mechanical property is still inadequate, which greatly limits the efficiency and success rate of injection. Thus, a new rate-dependent mechanical model based on membrane theory is proposed for the first time. In this model, an analytical equilibrium equation between the injection force and cell deformation is established by considering the speed effect of microinjection. Different from the traditional membrane-theory-based model, the elastic coefficient of the constitutive material in the proposed model is modified as a function of the injection velocity and acceleration, effectively simulating the influence of speeds on the mechanical responses and providing a more generalized and practical model. Using this model, other mechanical responses at different speeds can be also accurately predicted, including the distribution of membrane tension and stress and the deformed shape. To verify the validity of the model, numerical simulations and experiments are carried out. The results show that the proposed model can match the real mechanical responses well at different injection speeds.
\end{abstract}


\section{Introduction}
\label{sec_1}

\IEEEPARstart{M}{icroinjection} is an effective micromanipulation technique to deliver external materials (e.g., protein, DNA, and drug) into biological cells, and has been widely applied in biomedical fields, such as intracytoplasmic sperm injection (ICSI), drug discovery, and gene delivery \cite{zhang2018robotic,zhao2018review,permana2016review}. In microinjection, the injected cells are usually free-moving, deformable, and fragile, making rapid capture and successful injection difficult. To improve injection efficiency and success rate, lots of attentions have been paid to the development of microinjection techniques, including force sensing and control \cite{xie2010force,dai2019robotic,shang2022rotation}, visual servoing \cite{sakaki2009development,zhuang2017visual,fan2019precise}, batch injection \cite{huang2009robotic,lu2007micromanipulation,chi2022design}, haptic feedback and virtual training \cite{faroque2016haptic,ghanbari2012haptic}, etc. However, few works have been done on the mechanical properties of cells in microinjection. Due to the important role of cell mechanical properties in the regulation of life activities \cite{hao2020mechanical}, the limited cognition of them will conversely block the development of cell microinjection. Thus, it is essential to study the mechanical properties of cells. 

Cell mechanical property refers to the deformability when subject to the mechanical force, which has a significant impact on cell behaviors \cite{hao2020mechanical}. To characterize the cell mechanical properties, it is a feasible way to establish a mechanical model to simulate the relationship between force and deformation during microinjection \cite{tan2010characterizing}. Through the established mechanical model, the mechanical response information of cells can be predicted in advance, so as to better guide the cell micromanipulation. Due to the viscoelastic properties \cite{hao2020mechanical}, cells can exhibit both viscous and elastic properties, resulting in a complex nonlinear relationship between force and deformation, which makes it difficult to establish an accurate mechanical model. So far, the cell modeling methods mainly include continuum approach, micro/nano-structural approach and energetic approach \cite{lim2006mechanical,wei2018survey}. Compared with the other two methods, the continuum approach assumes that the cell is composed of elastic or viscoelastic continuum materials, and uses the appropriate constitutive materials to describe the deformation behavior of the cell. Since this kind of method is simple and intuitive, and the constitutive relation and related parameters of the material model can also be obtained from the experiment, it is often used in cell microinjection and has evolved several versions of mechanical models, like the point-load model \cite{sun2003mechanical}, the Maxwell-Wiechert model \cite{hajiyavand2019effect,chen2015speed}, and the Newtonian liquid drop model \cite{lim2006mechanical}.

Although the above studies have made good contributions, they can only provide limited information, such as injection force, cell deformation, and elastic modulus. For better understanding the cell mechanical response during microinjection, Tan \textit{et al}. \cite{tan2008mechanical, tan2010characterizing} firstly proposed a more detailed continuum mechanical model based on the membrane theory. The model can describe not only the injection force, the cell deformation, the membrane stress and tension distribution, the internal cell pressure and the deformed shape, but also the influences of injector radius, embryonic development stage, and constitutive materials. However, it requires the injection speed is constant over time. But in fact, the speed effect has a great influence on the mechanical properties, which cannot be ignored. Note that the so-called speed effect here includes the velocity and acceleration. As reported in \cite{liu2015analyses,hajiyavand2019effect,chen2015speed}, the injection speed affects the cell damage to a large extent, determining the efficiency and success rate of microinjection. Besides, the introduction of speed effect also generates complex dynamics problems, which makes the establishment of mechanical models a challenging task. To this end, this paper will devote to developing a cell mechanical model with considering speed effect on the basis of previous work \cite{tan2008mechanical} to obtain a more general and practical microinjection model. 
To the authors' knowledge, little relative research has been reported.

In this paper, a new continuum mechanical model is established based on the membrane theory for cell microinjection. In particular, 
an analytical equilibrium equation is first derived from the membrane theory to define the relationship between the injection force and cell deformation. Then, to describe the membrane behavior, the nonlinear Mooney-Rivlin constitutive material is adopted and its elastic coefficient is modified as a function of the injection velocity and acceleration instead of the traditional constant value \cite{tan2008mechanical,tan2010characterizing}. 
With the proposed model, some typical mechanical responses such as injection force, cell deformation, tension and stress distribution, and cell shape are accurately predicted at different speeds. To verify the effectiveness of the model, a series of simulations and zebrafish embryos microinjection experiments are carried out. The results show that the proposed model agrees well with the actual data and can be used to characterize the mechanical properties of cells under the different injection speeds.
To distinguish the difference from the traditional models without considering the speed effect, hereafter, the proposed model is regarded as ``rate-dependent'' mechanical model.

The remainder of this paper is organized as follows. Section \ref{sec_2} briefly introduces the membrane theory, and then the proposed improved model are derived in Section \ref{sec_3}. Afterwards, the simulation analysis and experimental verification are conducted in Sections \ref{sec_4} and \ref{sec_5}, respectively. Finally, Section \ref{sec_6} concludes this paper.

\section{Overview of the Membrane Theory}
\label{sec_2}

The membrane theory was firstly used to describe the cell mechanical model in microinjection by Tan \textit{et al}. \cite{tan2008mechanical}. It assumes that the biological cells are spherical and their biomembranes are composed of incompressible homogeneous isotropic material with constant thickness. The mechanical response generated by the intracellular liquid (i.e., cytoplasm) is considered as a uniform hydrostatic pressure on the biomembrane. So the deformation behavior of the cell is mainly determined by the external interaction force. Besides, since the membrane thickness is much smaller than the cell radius, the influence of the bending rigidity on the cell deformation is negligible comparing to that of the extensional rigidity.

\begin{figure}[!t]
	\centering
	\includegraphics[width=3.2in]{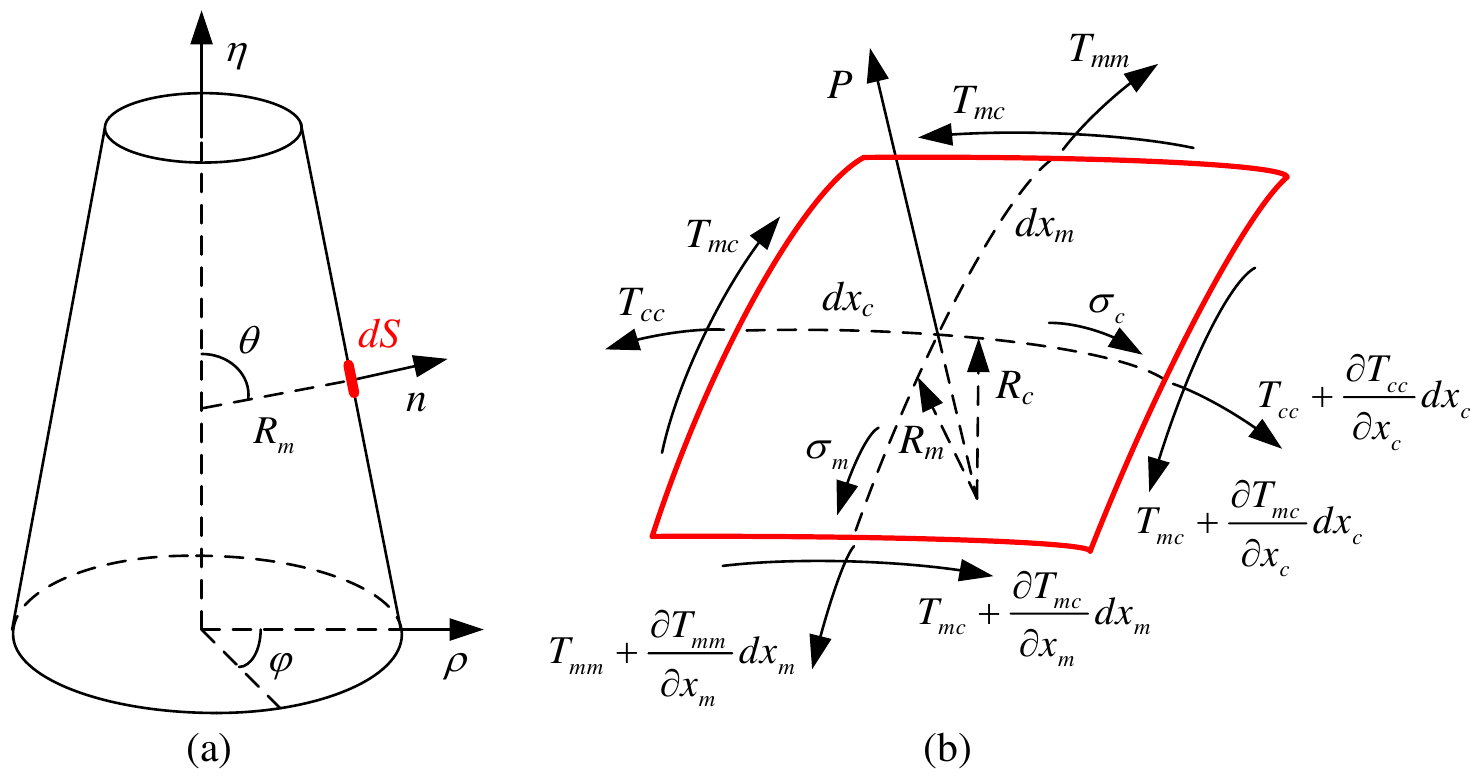}
	\caption{Schematic diagram of the rotationally symmetric membrane (a) and the force analysis of its differential element (b).}
	\label{fig_01}
\end{figure}

Based on the above assumptions, in order to employ the membrane theory to simulate the deformation of cells, without loss of generality, the mechanical equilibrium of a rotationally symmetric membrane is taken into account, as shown in Fig. \ref{fig_01}. Firstly, a small differential element is selected from the rotationally symmetric membrane in Fig. \ref{fig_01} (a), and its force analysis is drawn in Fig. \ref{fig_01} (b). For the sake of theoretical modeling, the shape of the membrane after deformation is defined by cylindrical coordinates $(\rho, \varphi, \eta)$, $x_m$ and $x_c$ are the principal axes of the membrane element, $dS$ is the arc differential along the meridian direction, $\theta$ is the angle between the normal $n$ and the axis $\eta$ at any point on the membrane surface, $R_m$ and $R_c$ are the principal curvature radii, $T_{mm}$ and $T_{cc}$ are the principal tensions, $T_{mc}$ is the shear force, $\sigma_m$ and $\sigma_c$ are the shear stresses in the directions of $x_m$ and $x_c$, and $P$ is the external pressure in the normal direction. To this end, according to the mechanical equilibrium condition of the membrane element that the resultant forces in the three independent orthogonal directions are all zero, one can obtain 
\begin{equation} \label{eq_01}
\sum{F_i} = 0 \quad (i = m,c,n)
\end{equation}
where the subscripts $m$, $c$ and $n$ represent the meridian, circumferential and normal directions of the membrane element, respectively.

\begin{figure}[!t]
	\centering
	\includegraphics[width=3.5in]{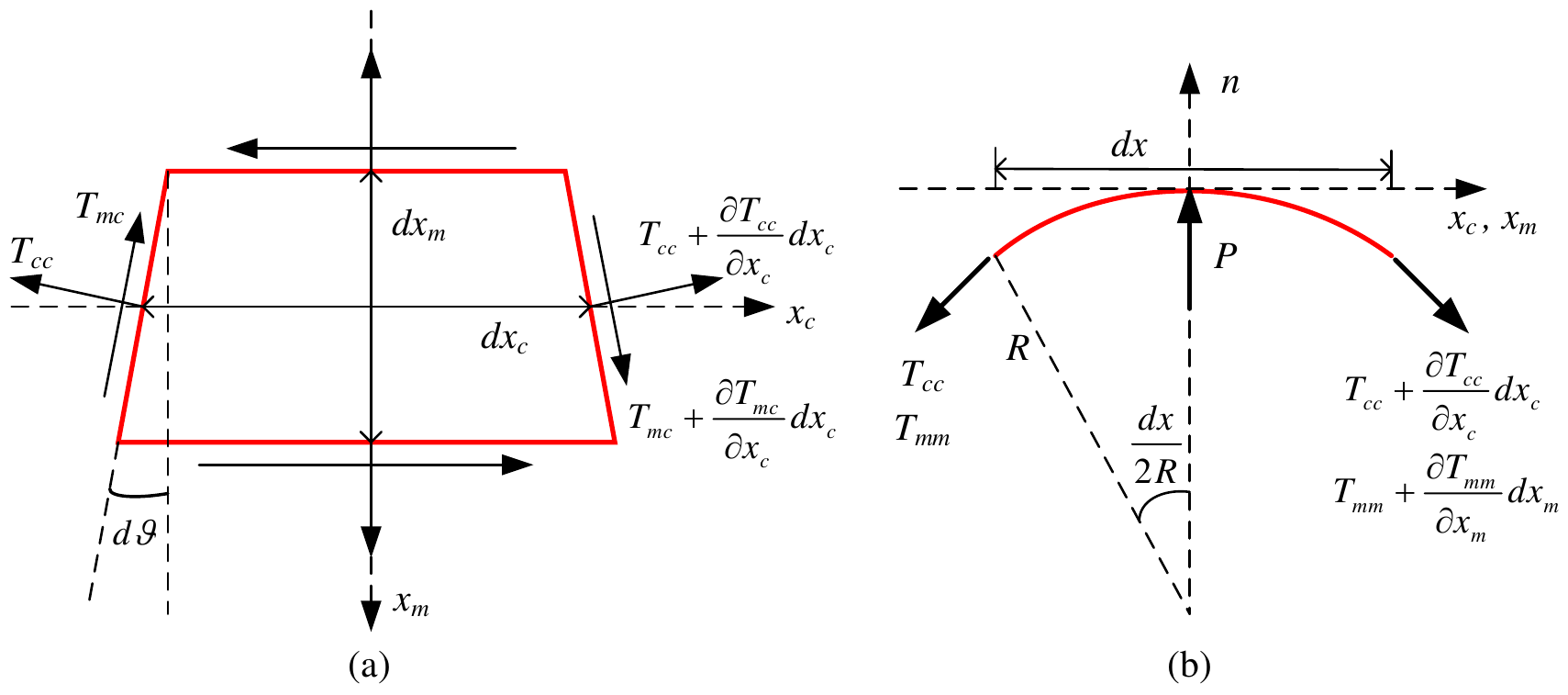}
	\caption{The projections of a differential membrane element onto the tangent plane (a) and the normal plane (b).}
	\label{fig_02}
\end{figure}

Thus, in the meridian direction $x_m$, the equilibrium equation is determined by
\begin{eqnarray} \label{eq_02}
\left( T_{mm} + \frac{\partial{T_{mm}}}{\partial{x_m}} d{x_m} \right) d{x_c} + \left( T_{mc} + \frac{\partial{T_{mc}}}{\partial{x_c}} d{x_c} \right) d{x_m}  \nonumber \\ 
+ \sigma_m d{x_c} d{x_m} - T_{mc} d{x_m} - T_{mm} d{x_c} + T_{sm} = 0
\end{eqnarray}
where $T_{sm}$ is produced by the shear effect at the boundary, and can be expressed as
\begin{equation} \label{eq_03}
T_{sm} =  - \left( T_{cc} + \frac{\partial{T_{cc}}}{\partial{x_c}} d{x_c} \right) d{x_m} d\vartheta - T_{cc} d{x_m} d\vartheta 
\end{equation}
where $d\vartheta $ is half of the included angle between the curved membrane element and its projection on the tangent plane (see Fig. \ref{fig_02} (a)), and satisfies $d\vartheta  = \frac{1}{2} \partial(d{x_c}) / \partial{x_m}$.

Substituting Eq. (\ref{eq_03}) into (\ref{eq_02}) yields
\begin{eqnarray} \label{eq_04}
\frac{\partial (T_{mm} d{x_c} )}{\partial{x_m}} d{x_m} + \frac{\partial{T_{mc}}}{\partial{x_c}} d{x_m} d{x_c} - T_{cc}\frac{\partial (d{x_c})}{\partial{x_m}} d{x_m}  \nonumber \\ 
 + \sigma_m d{x_m} d{x_c} = 0
\end{eqnarray}
Note that the multiplication of two identical differentials is almost infinitesimal and can be negligible, i.e., $d{x} \cdot d{x} \approx 0$.

Similar to the direction $x_m$, the equilibrium equation in the circumferential direction $x_c$ is obtained by
\begin{eqnarray} \label{eq_05}
\left( T_{cc} + \frac{\partial{T_{cc}}}{\partial{x_c}} d{x_c} \right) d{x_m} + \left( T_{mc} + \frac{\partial{T_{mc}}}{\partial{x_m}} d{x_m} \right) d{x_c}  \nonumber \\ 
+ \sigma_c d{x_c} d{x_m} - T_{mc} d{x_c} - T_{cc} d{x_m} + T_{sc} = 0
\end{eqnarray}
with ${T_{sc}} = ({T_{mc}} + \frac{{\partial {T_{mc}}}}{{\partial {x_c}}}d{x_c})d{x_m}d\vartheta  + {T_{mc}}d{x_m}d\vartheta $. Then, Eq. (\ref{eq_05}) can be simplified as
\begin{eqnarray} \label{eq_06}
\frac{\partial{T_{cc}}}{\partial{x_c}} d{x_c} d{x_m} + \frac{\partial (T_{mc} d{x_c}) }{\partial{x_m}} d{x_m} + T_{mc} \frac{\partial (d{x_c})}{\partial{x_m}} d{x_m}   \nonumber \\ 
+ \sigma_c d{x_c} d{x_m} = 0
\end{eqnarray}

Besides, in the normal direction of the membrane element, as shown in Fig. \ref{fig_02} (b), the equilibrium equation is expressed as
\begin{equation} \label{eq_07}
P d{x_c} d{x_m} - T_{mm} d{x_c} \frac{d{x_m}}{R_m} - T_{cc} d{x_m} \frac{d{x_c}}{R_c} = 0
\end{equation}

Combining with the relationships $d{x_c} = \rho d\varphi $ and $d{x_m} = dS$, Eqs. (\ref{eq_04}), (\ref{eq_06}) and (\ref{eq_07}) are further rewritten as
\begin{eqnarray}
\frac{\partial (T_{mm} \rho )}{\partial S} + \frac{\partial T_{mc}}{\partial \varphi } - T_{cc} \frac{\partial \rho }{\partial S} + \sigma_m \rho  = 0   \label{eq_08}  \\ 
\frac{\partial T_{cc}}{\partial \varphi } + \frac{\partial (T_{mc} \rho )}{\partial S} + T_{mc} \frac{\partial \rho}{\partial S} + \sigma_c \rho  = 0   \label{eq_09}  \\ 
\frac{T_{mm}}{R_m} + \frac{T_{cc}}{R_c} = P  \label{eq_010}
\end{eqnarray}

Based on Eqs. (\ref{eq_08})-(\ref{eq_010}), the theoretical foundation is built for cell mechanical model. Next, one will introduce the proposed improved method based on the membrane theory to model biological cells with considering speed effect of microinjection.

\section{Improved Rate-Dependent Mechanical Model}
\label{sec_3}

\subsection{Model Derivation}

For the traditional mechanical model proposed by Tan \textit{et al}. \cite{tan2008mechanical}, it exhibits rate-independently, i.e., it does not consider the effects of velocity and acceleration of microinjection. To this end, on the basis of the previous work \cite{tan2008mechanical}, this paper proposes an improved rate-dependent model to describe the mechanical properties of cell microinjection.

Generally, biological cells are divided into two types: suspended cells and adherent cells. In this paper, the developed mechanical model is devoted to rotationally symmetric suspended cells. During microinjection, an external force is usually imposed directly towards the center of cell. For such axisymmetric loading case, the shear stress $\sigma_c$ and $T_{mc}$ are both zero, then the equilibrium equations (\ref{eq_08})-(\ref{eq_010}) can be simplified as
\begin{eqnarray}
\frac{\partial (T_m \rho)}{\partial S} - T_c \frac{\partial \rho }{\partial S} + \sigma_m \rho  = 0   \label{eq_011}  \\ 
K_m T_m + K_c T_c = P  \label{eq_012}
\end{eqnarray}
where $T_m$ and $T_c$ now represent the principal tensions, $K_m$ and $K_c$ are the principal curvatures.

\begin{figure}[!t]
	\centering
	\includegraphics[width=3.2in]{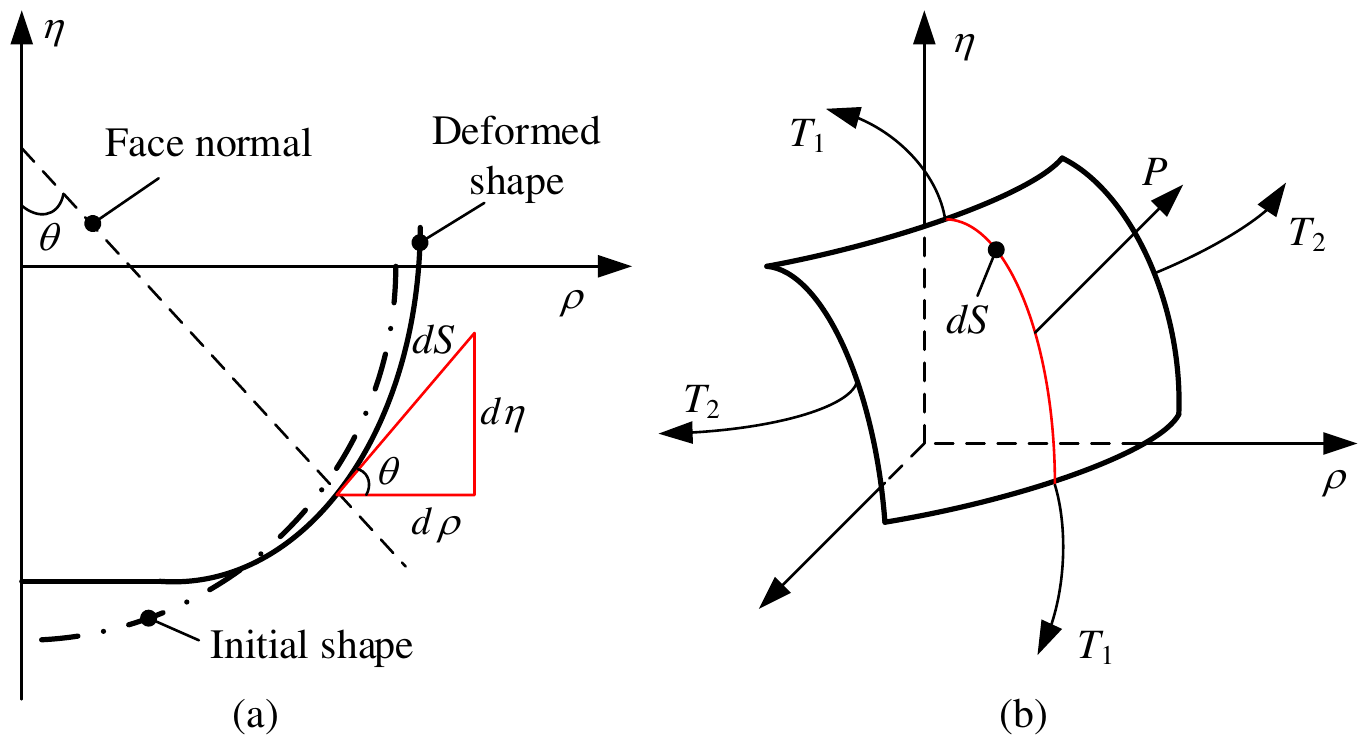}
	\caption{Differential geometric relations of cell membrane after deformation (a) and the force diagram (b).}
	\label{fig_03}
\end{figure}

Using Eqs. (\ref{eq_011}) and (\ref{eq_012}), the mechanical equilibrium of rotationally symmetric membrane can be simulated. Assuming that the thickness of cell membrane is $h$ and the initial radius is $r_0$, one can use the spherical coordinates $(r,\Theta ,\psi )$ to describe the cell membrane before deformation. When the external force is applied to the center of the cell, the spherical membrane deforms but still retains rotational symmetry. Thus, the cylindrical coordinates $(\rho ,\Theta ,\eta )$ are used to describe the shape of the deformed membrane. Accordingly, the arc differential $ds$ in the meridian direction of the membrane before deformation becomes $dS$ after deformation, as depicted in Fig. \ref{fig_03}. Similarly, the arc differential in the circumferential direction is also changed from $r_0 \sin \psi d \Theta$ to $\rho d \Theta$. 
Therefore, from the coordinate definitions of the initial shape and deformed shape, the principal stretch ratios $\lambda_m$ and $\lambda_c$ can be expressed as
\begin{eqnarray}
\lambda_m = \frac{dS}{ds} = \frac{\sqrt{{d \rho}^2 + {d \eta}^2}}{r_0 d \psi} = \frac{\sqrt{{\rho'}^2 + {\eta'}^2}}{r_0}  \label{eq_013}  \\ 
\lambda_c = \frac{\rho d \Theta}{r_0 \sin \psi d \Theta} = \frac{\rho}{r_0 \sin \psi } \label{eq_014}
\end{eqnarray}

And in view of the geometric relationship between $dS$, $d \rho$ and $d \eta$ in Fig. \ref{fig_03} (a), one also get
\begin{equation} \label{eq_015}
\cos \theta  =  \pm \frac{d \rho}{dS} = \pm \frac{\rho'}{\sqrt {{\rho'}^2 + {\eta'}^2}} = \pm \frac{(\lambda_c \sin \psi )'}{\lambda_m}  
\end{equation}
which further derives that $\theta = \arccos \left[ \pm (\lambda_c \sin \psi )' / \lambda_m \right] $. Note that the symbol ``$ \pm $'' here should be selected to ensure that the corresponding value is positive. And the prime $(\cdot)'$ stands for the derivative with respect to the angle $\psi$ in the foregoing and subsequent equations.

Thus, the principal curvatures $K_m$ and $K_c$ in Eq. (\ref{eq_012}) are deduced from Eqs. (\ref{eq_013})-(\ref{eq_015}) as
\begin{eqnarray}
K_m =  \pm \frac{d \theta}{dS} =  \pm \frac{\lambda_m' (\lambda_c \sin \psi )' - (\lambda_c \sin \psi)'' \lambda_m}{r_0 \lambda_m^3 \sqrt{ 1 - \frac{{(\lambda_c \sin \psi )'}^2}{\lambda_m^2}}}   \label{eq_016}  \\  
K_c = \frac{\sin \theta}{\rho} = \frac{\sqrt{\lambda_m^2 - {(\lambda_c \sin \psi)'}^2}}{r_0 \lambda_m \lambda_c \sin \psi}  \label{eq_017}
\end{eqnarray}
And Eq. (\ref{eq_011}) can be also rewritten as
\begin{equation} \label{eq_018}
\frac{\partial{T_m}}{\partial{\lambda _m}} \lambda_m' + \frac{\partial {T_m}}{\partial{\lambda_c}} \lambda_c' = \frac{\rho'}{\rho}(T_c - T_m) - \frac{\sigma_m \rho'}{\cos \theta}
\end{equation}
See Appendix \ref{app_1} for the deducing process of Eq. (\ref{eq_018}) in detail. 

\begin{figure}[!t]
	\centering
	\includegraphics[width=2.8in]{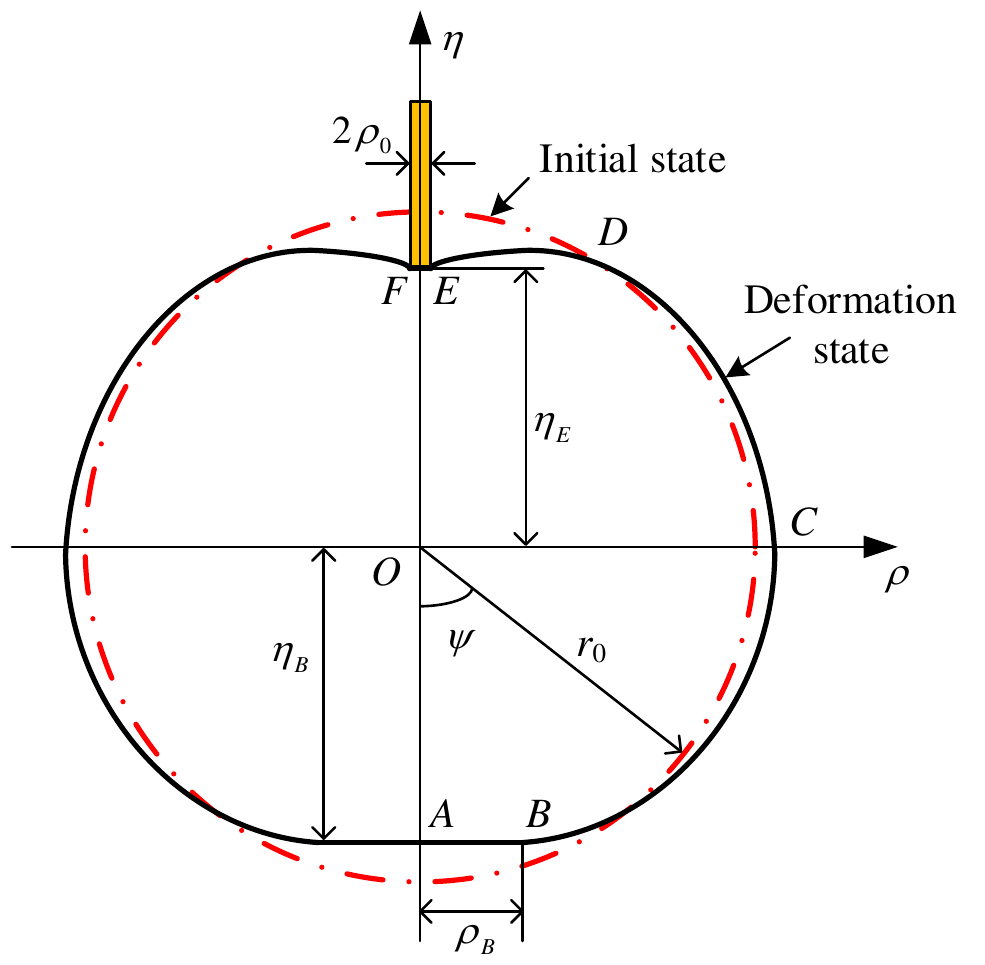}
	\caption{Coordinates definition of the cell membrane for different states \cite{tan2008mechanical}.}
	\label{fig_04}
\end{figure}

Fig. \ref{fig_04} illustrates the coordinates definition of cell membrane before and after microinjection. As mentioned before, the force applied to the cell and the resulting deformation are symmetric during microinjection. Therefore, only half of the deformed cells need to be analyzed. As seen from Fig. \ref{fig_04}, there are two different areas: the flat area, e.g., $AB$ and $EF$, and the bending area, e.g., $BC$, $CD$ and $DE$. 

For the flat area, the following conditions needs to hold:
\begin{equation} \label{eq_019}
\eta ' = 0, \, d \theta = 0, \, K_m=K_c=0
\end{equation}
Therefore, according to Eqs. (\ref{eq_013}), (\ref{eq_014}), and (\ref{eq_018}), the governing equations are obtained by
\begin{eqnarray}
\lambda_m' =  \pm \frac{\lambda_m f_3}{\lambda_c f_1 \sin \psi} - \frac{ (\pm \lambda_m - \lambda_c \cos \psi) f_2}{f_1 \sin \psi} \mp \frac{\sigma_m r_0 \lambda_m}{f_1}   \label{eq_020} \\  
\lambda_c'  = \frac{ \pm \lambda_m - \lambda_c \cos \psi }{\sin \psi }  \label{eq_021}
\end{eqnarray}
where ``$ + $'' is for $AB$ area, and ``$ - $'' is for $EF$ area. 

For the bending area, substituting Eqs. (\ref{eq_016}) and (\ref{eq_017}) into (\ref{eq_012}) and together with Eq. (\ref{eq_018}), one can obtain the governing equations as
\begin{eqnarray}
\lambda_m' = \frac{(\delta \cos \psi - \omega \sin \psi) f_2 }{f_1 \sin ^2 \psi } + \frac{\omega f_3}{\delta f_1} \mp \frac{\sigma_m r_0 \lambda_m}{f_1}   \label{eq_022} \\  
\omega ' = \frac{\lambda_m' \omega }{\lambda_m} \pm \left[ \frac{(\lambda_m^2 - \omega ^2) T_c}{\delta T_m} - \frac{\lambda_m P r_0 \sqrt{\lambda_m^2 - \omega ^2} }{T_m}\right]  \label{eq_023}
\end{eqnarray}
with $\delta = \lambda_c \sin \psi $, $\omega = \delta '$, $f_1 = \partial T_m / \partial \lambda _m$, $f_2 = \partial T_m / \partial \lambda_c$ and $f_3 = T_c - T_m$. Here, ``$ + $'' is for $BC$ and $CD$ areas, and ``$ - $'' is for $DE$ area. 
%The deducing processes of Eqs. (\ref{eq_022})-(\ref{eq_023}) in detail are given in Appendix \ref{app_2}.

For modeling the membrane behaviors of cells, constitutive materials are used to define the relations between the membrane tensions and strains. In the traditional rate-independent mechanical model \cite{tan2008mechanical}, the constitutive material of cell membrane is selected as the Mooney-Rivlin hyperelastic material thanks to the benefit of physically nonlinear and incompressible characteristics. The corresponding strain energy function takes the following form:
\begin{equation}  \label{eq_024}
W = C_1 (I_1 - 3) + C_2 (I_2 - 3) = C_1 [(I_1 - 3) + \alpha (I_2 - 3)]
\end{equation}
where $C_1$ and $C_2$ are the elastic coefficients, respectively, and the unit is consistent with the stress unit and $\alpha=\frac{C_2}{C_1}$. For homogeneous, isotropic and incompressible elastic materials, $C_1 = \frac{E}{6(1 + \alpha )}$, where $E$ is the elastic modulus. $I_1$ and $I_2$ are strain invariants and can be expressed as functions of the principal stretch ratios $\lambda_m$ and $\lambda_c$ in Eqs. (\ref{eq_013})-(\ref{eq_014}) by
\begin{eqnarray}
I_1 = \lambda_m^2 + \lambda_c^2 + \frac{1}{\lambda_m^2 \lambda_c^2}   \label{eq_025}  \\ 
I_2 = \lambda_m^2 \lambda_c^2 + \frac{1}{\lambda_m^2} + \frac{1}{\lambda _c^2}  \label{eq_026}
\end{eqnarray}

In view of Eqs. (\ref{eq_024}), one can find that the elastic coefficient $C_1$ of the defined constitutive material is a constant. Although this kind of material simulates the membrane behaviors well under a constant speed, once the injection speed changes, the value of $C_1$ needs to be re-identified accordingly due to the inherent variable stiffness characteristic of cells \cite{hao2020mechanical}. Otherwise, the prediction results via the theoretical model will go away from the reality, which has been proved in the subsequent simulation and experimental results. In other words, the cell mechanical model defined by the constitutive material (\ref{eq_024}) is a static rate-independent model. Thus, in order to determine the effect of injection velocities and accelerations on cell deformation, the rate-dependent constitutive material model is further proposed to eliminate the limitation of the traditional cell mechanical model \cite{tan2008mechanical}.

According to the description in \cite{mohotti2014strain}, the strain rate-dependent behavior of hyperelastic constitutive materials, such as the Mooney-Rivlin material, can be described in combination with the strain energy absorption capacity of the materials. From Eq. (\ref{eq_024}), it is found that principal stretch ratios and elastic coefficients are key factors affecting the strain energy, which correspond to the cell deformation and the rigidity of cell membrane, respectively. Considering the fact that the rigidity of cell membrane changes significantly with injection speeds, the elastic coefficient $C_1$ is thus modified to be a rate-dependent term as follows:
\begin{equation} \label{eq_027}
C_1 = (k_2 v^2 + k_1 v + k_0)(g_0 + \frac{1}{g_1 + g_2 a})
\end{equation}
where $v$ and $a$ are the velocity and acceleration of injection, $k_i$ and $g_i$ are the coefficients to be identified. For a constant injection velocity $v$, the value $a$ is zero; while for an accelerated injection with $a$, the value $v$ denotes the instantaneous velocity of puncturing the cell membrane. Besides, in the right side of Eq. (\ref{eq_027}), the first term $(k_2 v^2 + k_1 v + k_0)$ describes the situation under the constant injection velocity, denoted as $C_{1v}$. Since the value of $C_{1v}$ tends to reduce under the accelerated injection (see the experimental results in Section \ref{sec_5_2}), the second term $(g_0 + \frac{1}{g_1 + g_2 a})$ is introduced to reflect the degree of reduction, denoted as $\varepsilon$. Under the actions of $C_{1v}$ and $\varepsilon$, the rate-dependent property of the proposed mechanical model is finally guaranteed.

\begin{remark}
It is worth noting that for the traditional cell model \cite{tan2008mechanical}, the coefficient $C_1$ is a constant value, which does not consider the speed effect of microinjection. But in the proposed model, the velocity and acceleration of injection are all taken into account, leading to a more generalized and practical mechanical model. The superiority will be demonstrated in Sections \ref{sec_4} and \ref{sec_5}.
\end{remark}

Based on the strain energy function, the principal stresses and principal tensions of cell membrane can be derived as follows:
\begin{eqnarray}
\sigma_i = \lambda_i \frac{\partial W}{\partial \lambda_i}, \; (i = m, c)   \label{eq_028}  \\ 
T_i = \frac{h}{\lambda_m \lambda_c} \sigma_i, \; (i = m, c) \label{eq_029}
\end{eqnarray}

Moreover, the cell deformation and the injection force are also obtained by
\begin{eqnarray}
d = 2 r_0 - (\eta_E - \eta_B)   \label{eq_030}  \\ 
F = P \pi \rho_B^2 = P \pi (r_0 \lambda_{cB} \sin \psi_B)^2 \label{eq_031}
\end{eqnarray}
where $\eta_B$ and $\eta_E$ are calculated from Eqs. (\ref{eq_013}) and (\ref{eq_014}), and satisfy that $\eta'= \pm r_0 \sqrt{\lambda_m^2 - (\lambda_c \sin \psi)'^2}$, ``$ + $'' is used for $\eta_B$, and ``$ - $'' is for $\eta_E$.

It can be found from Eqs. (\ref{eq_028})-(\ref{eq_031}), the injection force $F$, the cell deformation $d$, the membrane tensions $T_i$ and stresses $\sigma_i$ are all related to the principle stretch ratios $\lambda_m$, $\lambda_c$ and the specified angle $\psi$. Moreover, if an initial angle $\psi_B$ is prescribed first, the principle stretch ratios $\lambda_m$ and $\lambda_c$ are calculated with respect to $\psi$ through governing equations (\ref{eq_020})-(\ref{eq_023}). Then, the relationship between the injection force and the deformation, the mechanical responses of membrane stress and tension distribution, and the deformed cell shape can be obtained accordingly.

\subsection{Solution Algorithm}

\begin{algorithm}[!t]  
	\caption{The solution algorithm based on the proposed rate-dependent cell mechanical model.}  
	\label{algo_1}
	\begin{algorithmic}[1]  
		\State Set the constants $\left\lbrace r_0, h, \rho_0, \alpha \right\rbrace $, and calculate the initial volume $V_0$ by Eq. (\ref{eq_041});
		\State Initialize the parameters $\left\lbrace v, a, k_i, g_i \right\rbrace $, and calculate $C_1$ by Eq. (\ref{eq_027});  
		\State Prescribe an angle $\psi_B$ to determine the cell deformation;
		\While {$V \neq V_0$} 
		\State Assume a principal stretch ratio $\lambda_A$;
		\State Solve Eqs. (\ref{eq_020}), (\ref{eq_021}) for $AB$ area by the Runge-Kutta method under the boundary condition (\ref{eq_032});		 
		\While {$\omega \neq 0$}  
		\State Assume a pressure $P$;
		\State Solve Eqs. (\ref{eq_022})-(\ref{eq_023}) for $BC$ area by the Runge-Kutta method under the boundary conditions (\ref{eq_033})-(\ref{eq_034});	    
		\EndWhile
		\While {$\omega \neq -\lambda_m$}  
		\State Assume a value $\psi_D$;
		\State Solve Eqs. (\ref{eq_022})-(\ref{eq_023}) for $CD$ area by the Runge-Kutta method under the boundary conditions (\ref{eq_035})-(\ref{eq_036});	    
		\EndWhile
		\While {$\rho_E \neq \rho_0$}  
		\State Assume a value $\psi_E$;
		\State Solve Eqs. (\ref{eq_022})-(\ref{eq_023}) for $DE$ area by the Runge-Kutta method under the boundary conditions (\ref{eq_037})-(\ref{eq_038});	    
		\EndWhile						
		\Repeat  
		\State Assume a value $\lambda_F$;
		\State Solve Eqs. (\ref{eq_020}), (\ref{eq_021}) for $EF$ area by the Runge-Kutta method under the boundary condition (\ref{eq_040});	    
		\Until {$\rho_E = \rho_0$} 
		\State Calculate the cell volume $V$ by Eq. (\ref{eq_042});			
		\EndWhile
		\State Output the deformed cell shape $\left\lbrace \rho, \eta \right\rbrace$  by solving Eqs. (\ref{eq_013})-(\ref{eq_014}), the principal stresses $\sigma_i$ and tensions $T_i$ by Eqs. (\ref{eq_028})-(\ref{eq_029}), the cell deformation $d$ and injection force $F$ by Eqs. (\ref{eq_030})-(\ref{eq_031});        
	\end{algorithmic}  
\end{algorithm} 

Since the established governing equations for both of the flat and bending areas are all ordinary differential equations (ODEs) related to the angle $\psi$, numerical solutions can be used to solve them, such as the classical fourth-order Runge-Kutta method \cite{tan2008mechanical}. To implement the solution algorithm, the boundary conditions for each ODE are
\begin{eqnarray}
&& \psi= \psi_A, \  \lambda_m=\lambda_c=\lambda_A  \label{eq_032}  \\
&& \psi = \psi_B, \  \omega = \lambda_m  \label{eq_033}  \\
&& \psi = \psi_B, \  (\lambda_m)_{BC} = (\lambda_m)_{AB}, (\delta)_{BC} = (\delta)_{AB}  \label{eq_034}  \\
&& \psi = \psi_C, \  \omega = 0  \label{eq_035}  \\ 
&& \psi = \psi_C, \  (\lambda_m)_{CD} = (\lambda_m)_{BC}, (\delta)_{CD} = (\delta)_{BC}  \label{eq_036}  \\ 
&& \psi = \psi_D, \  \omega = - \lambda_m   \label{eq_037}  \\
&& \psi = \psi_D, \  (\lambda_m)_{DE} = (\lambda_m)_{CD}, (\delta)_{DE} = (\delta)_{CD}   \label{eq_038}  \\
&& \psi = \psi_E, \  \rho_E = \rho_0    \label{eq_039}   \\
&& \psi= \psi_F, \  \lambda_m=\lambda_c=\lambda_F  \label{eq_040}
\end{eqnarray}
where $\lambda_{A,F}$ denote the stretch ratios of the points $A$ and $F$, $\rho_0$ is the radius of the injection needle, and $\psi_{A \rightarrow F}$ are the angles corresponding to the turning points described in Fig. \ref{fig_04}.

According to the assumption of the membrane theory at the beginning of Section \ref{sec_2} that biological cells are incompressible, therefore, the cell volume keeps constant during microinjection. The initial volume before injection is 
\begin{equation} \label{eq_041}
V_0 = \frac{4}{3} \pi r_0^3
\end{equation}
and the volume after injection is 
\begin{eqnarray}  \label{eq_042}
V = \pi r_0^3 \Big( \int_{\psi_B}^{\psi_C} \nu d \psi + \int_{\psi_C}^{\psi_D} \nu d \psi - \int_{\psi_D}^{\psi_E} \nu d \psi \Big) 
\end{eqnarray}
with $\nu = \delta^2 \sqrt{\lambda_m^2 - \delta'^2} $. So the constraint condition for the cell volume is $V = V_0$.

Based on the above boundary and constraint conditions, the governing equations (\ref{eq_020})-(\ref{eq_023}) can be solved, and the corresponding solution algorithm is given in Algorithm \ref{algo_1}. It should be mentioned that, for each prescribed $\psi_B$, when the parameters $\left\lbrace \lambda_A, P, \psi_D, \psi_E, \lambda_F \right\rbrace $ satisfy the specified conditions, the principle stretch ratios $\lambda_m$ and $\lambda_c$ are easily obtained in the flat and the bending areas. Combining with Eqs. (\ref{eq_013})-(\ref{eq_014}) and (\ref{eq_028})-(\ref{eq_029}), the deformed cell shape $\left\lbrace \rho, \eta \right\rbrace $, the membrane stresses $\sigma_i$ and the tensions $T_i$ can be further calculated, respectively. After obtaining $\left\lbrace \rho, \eta \right\rbrace $, the injection force $F$ and the cell deformation $d$ are accordingly provided via Eqs. (\ref{eq_030})-(\ref{eq_031}).  

\section{Simulation Analysis}
\label{sec_4}

\begin{figure}[!t]
	\centering
	\includegraphics[width=3.5in]{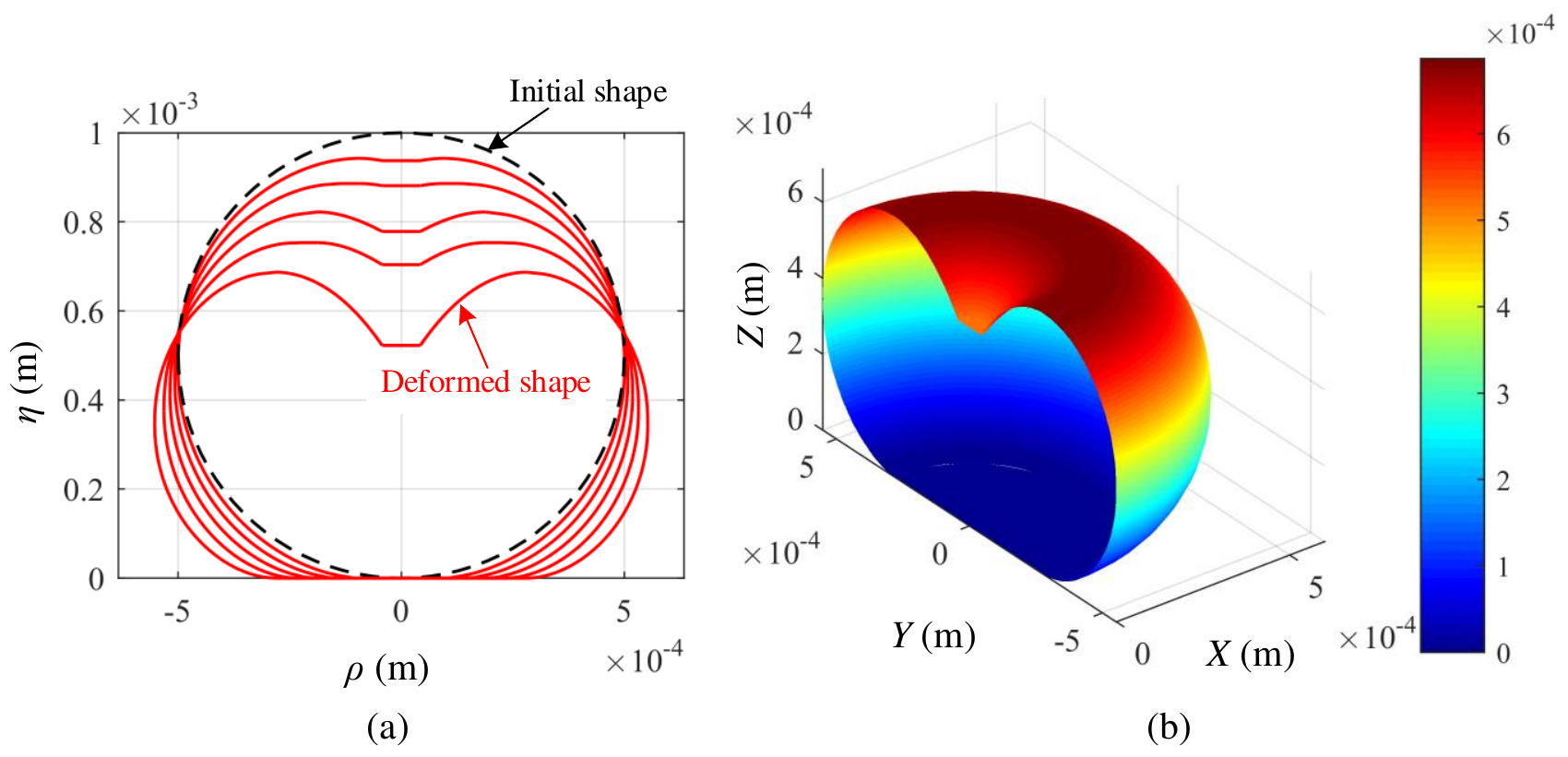}
	\caption{(a) The shapes of the deformed cell membrane obtained by the proposed rate-dependent mechanical model with different prescribed $\psi_B$ at an injection velocity $v=1.0$ mm/s. (b) 3D view of half of the deformed cell.}
	\label{fig_05}
\end{figure}

\begin{figure}[!t]
	\centering
	\includegraphics[width=3.5in]{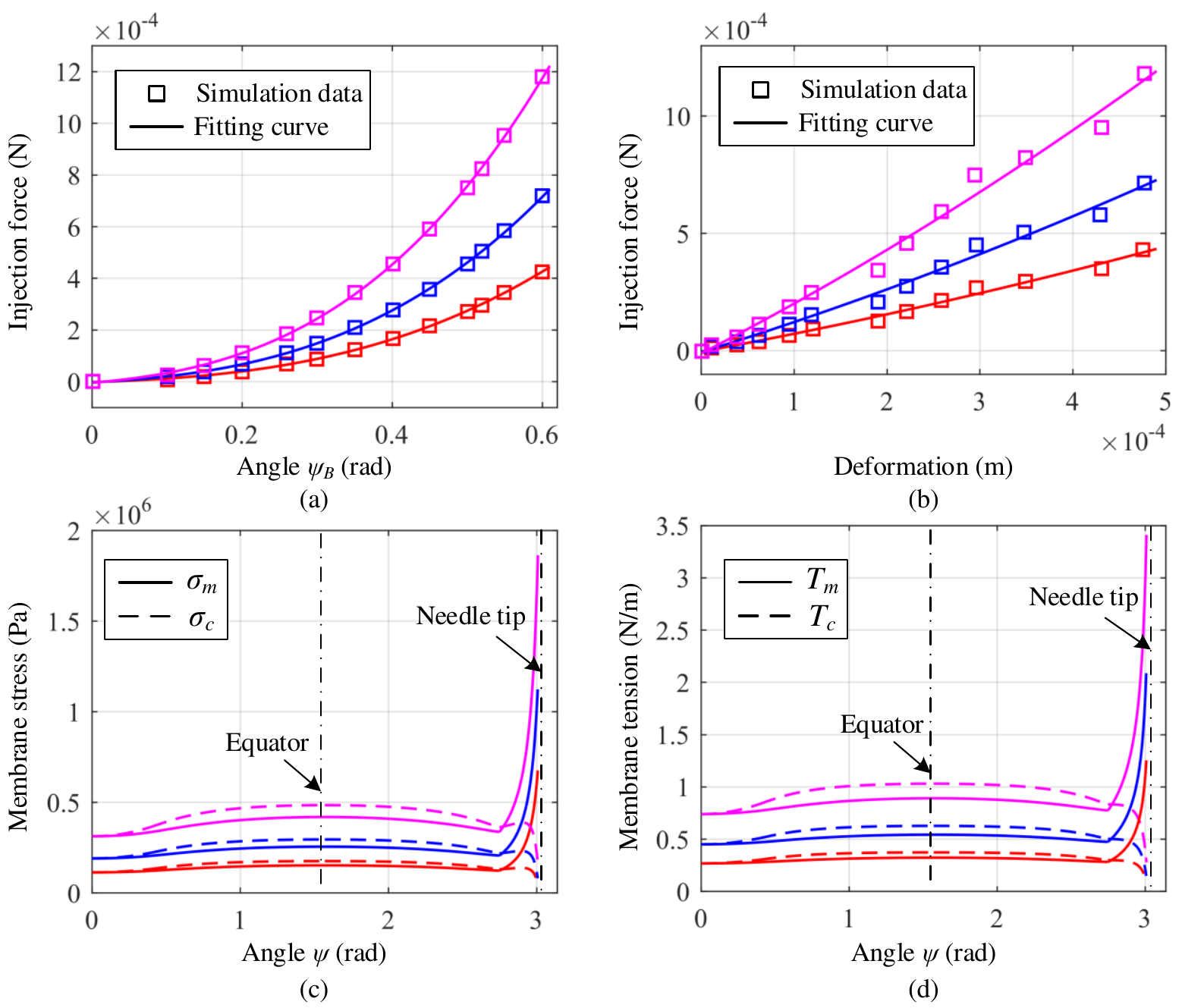}
	\caption{Simulation results using the proposed rate-dependent mechanical model under different injection velocities $v$. (a) The injection force vs. the angle $\psi_B$. (b) The injection force vs. the deformation of cell. (c) Membrane stress distribution with $\psi_B = 0.4$ rad. (d) Membrane tension distribution with $\psi_B = 0.4$ rad. Besides, in this figure, the red color stands for the case of injection velocity $v=0.2$ mm/s, the blue color for $v=0.6$ mm/s, and the magenta color for $v=1.0$ mm/s.}
	\label{fig_06}
\end{figure}

In order to analyze the cell mechanical properties characterized by the proposed rate-dependent model, the essential simulation parameters are selected as follows:$\alpha = 0.2$, $C_1 = (0.0875 v^2+0.0495v+0.057)(0.6066+\frac{1}{2.5333+3.3922a}) \, \text{ MPa}$, $h = 3 \, \mu \text{m}$, $r_0 = 500 \, \mu \text{m}$, $\rho_0 = 40 \, \mu \text{m}$. The solution algorithm is implemented by Matlab software (version R2021b). Figs. \ref{fig_05} and \ref{fig_06} show the numerical simulation results of cell mechanical properties under the different injection velocities.

It needs to mention that the radii of injection needle and the developmental stages of embryos still play important roles on the cell force and deformation, which have been discussed in \cite{tan2008mechanical}. Besides, different constitutive materials have been also studied and compared in \cite{tan2010characterizing}. To this end, in this work, they are not discussed repeatedly any more, and only the speed effect of microinjection is selected as the topic to make comparisons.

Fig. \ref{fig_05} shows the deformation process of the cell membrane during microinjection. It is found that, with the increase of the injection needle feed, the angle $\psi_B$ also increases, that is, the flat area of the cell bottom becomes larger. Therefore, the angle $\psi_B$ in the mechanical model determines the size of the deformed cell shape. Moreover, under the combined action of the external injection force and the internal pressure, the cell membrane continues to deform until rupture. From this result, one can predict the cell deformation shape well in the practical microinjection.

To illustrate the influence of injection speeds on the mechanical properties, Fig. \ref{fig_06} demonstrates the mechanical response curves of the injection force and deformation, the membrane stress and tension distribution under different injection velocities. 
It is seen from Fig. \ref{fig_06} (a) and (b) that, with the increase of injection velocity, the injection force increases dramatically under the same cell deformation condition, and the slope of the force-deformation curve gets steeper. Since the slope reflects the elastic modulus of the cell membrane, the injection velocity is positively related to the elastic coefficient of the hyperelastic material. This fully confirms the rationality of rate-dependent improvement on the coefficient $C_1$ in Eq. (\ref{eq_027}).
As a result, when selecting a larger injection speed, a smaller deformation will be generated with the same injection force, which is greatly beneficial for the survival and success rates of injected cells, simultaneously increases the microinjection efficiency.

Besides, from Fig. \ref{fig_06} (c) and (d), it can be observed that, under the different injection velocities, although the trends of the membrane tension and stress curves are basically the same, their distribution are not uniform, and there are two extreme values locating on the equator and near the needle tip. The magnitudes of both the tension and stress near the needle tip are growing very quickly, and are much bigger than those on the equator, which indicates that the rupture location of the cell membrane will occur near the tip. This phenomenon is consistent with the fact.

From the numerical simulation, it shows that, when given an initial angle $\psi_B$, the relationship between the injection force and the deformation of cell, and other mechanical properties like the membrane tension and the stress distribution, can be all determined by the proposed model. Besides, compared with the traditional membrane-theory-based model \cite{tan2008mechanical}, the proposed rate-dependent model not only inherits the advantages, but also consider the speed effect of microinjection, which effectively extend the application ranges, including the situations on the different radii of injection needle, the different constitutive materials, and the different developmental stages considered in \cite{tan2008mechanical,tan2010characterizing}.

\section{Experimental Verification}
\label{sec_5}
\subsection{Experimental Setup}

\begin{figure}[!t]
	\centering
	\includegraphics[width=3.5in]{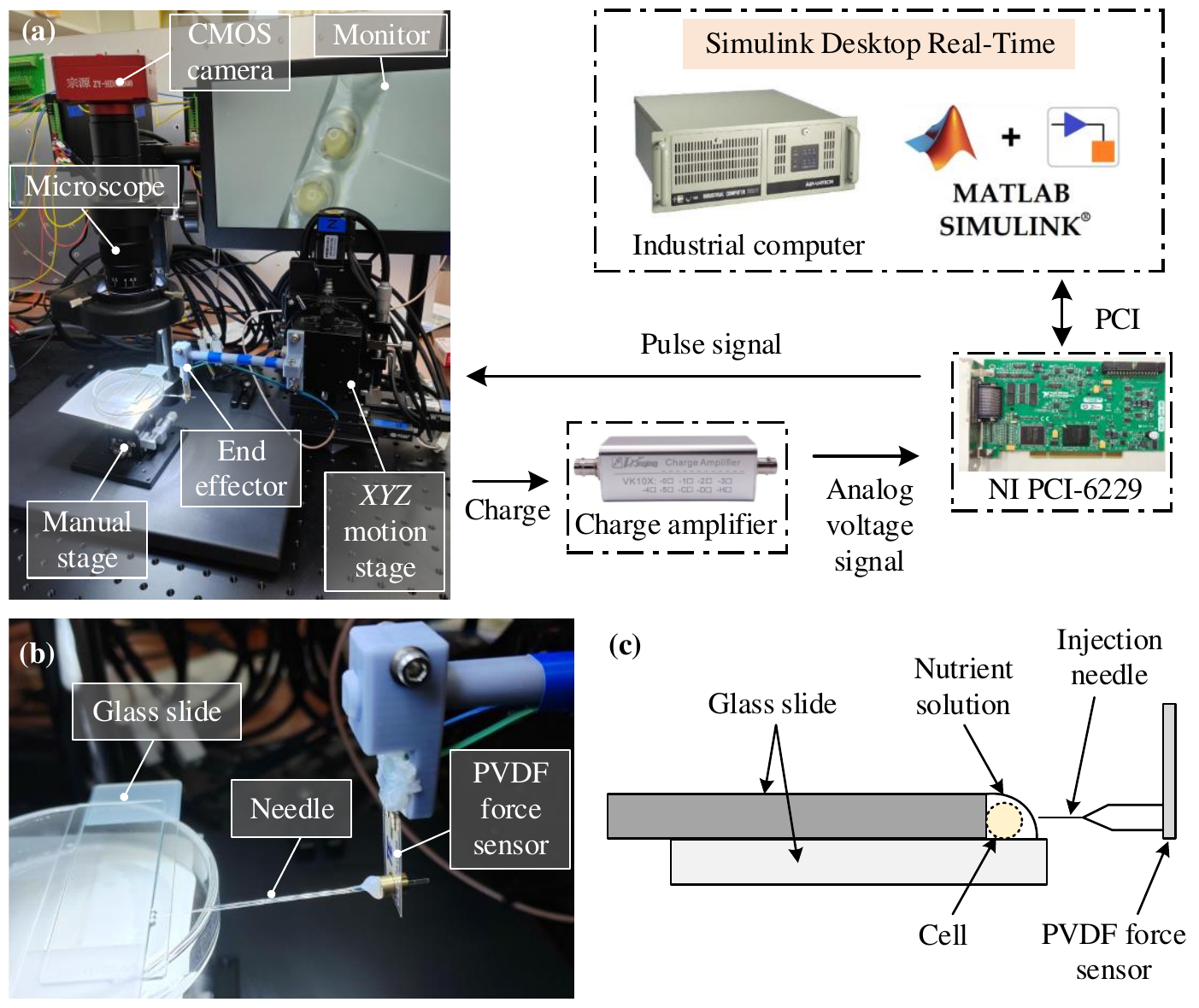}
	\caption{(a) Experimental setup for cell microinjection. (b) Structure of the end effector. (c) Schematic diagram of cell injection force measurement.}
	\label{fig_07}
\end{figure}

To verify the effectiveness of the proposed cell mechanical model, the experimental setup for cell microinjection is built as illustrated in Fig. \ref{fig_07}. It is mainly composed of the following parts: an $XYZ$ motion stage, an end effector, a manual stage, a vision imaging system, and a motion control system. The $XYZ$ motion stage is driven by three DC stepping motors with embedded encoders to achieve the closed-loop position control. Each axis has the motion range of 30 mm and the accuracy of 0.25 $\mu$m. The end effector includes a needle with the tip radius of 50 $\mu$m and a cantilever-type PVDF force sensor, as shown in Fig. \ref{fig_07} (b). It is mounted on the $XYZ$ motion stage to inject the cell and measure the injection force at the same time. Charge signals produced by the PVDF sensor are converted into voltage signals via a charge amplifier. Cells are placed and immobilized on the glass slides by the means in Fig. \ref{fig_07} (c). The manual stage with three translational movements is used to provide assisted adjusting for the relative position between the cells and the needle, such that the tip of needle is aligned with the center of the injected cells. The vision imaging system consists of an optical microscope, a CMOS camera, and a monitor to capture the cell injection motions. The motion control system contains an industrial computer and a data acquisition card (NI PCI-6229) to produce the control pulse signals for the three motors and acquire the analog voltage signal from the charge amplifier. The real-time control algorithm is implemented via the Desktop Real-Time environment in Matlab/Simulink software.

\begin{figure}[!t]
	\centering
	\includegraphics[width=3.2in]{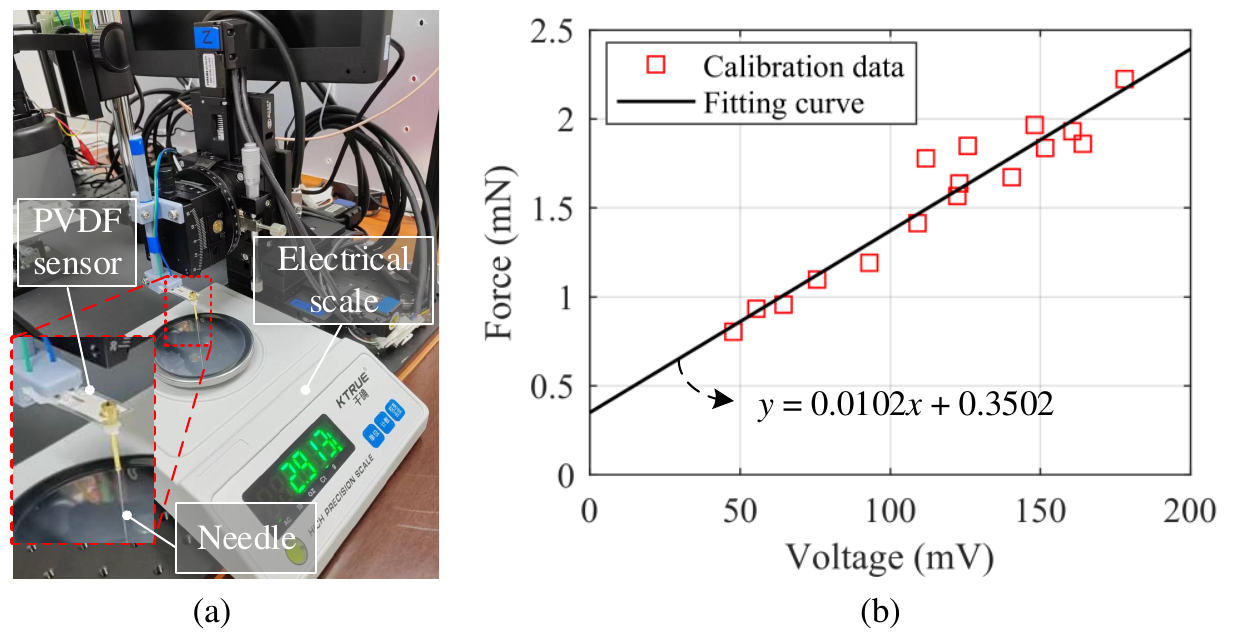}
	\caption{Calibration setup (a) and the corresponding result (b) of the PVDF force sensor. Note that the needle is perpendicular to the electrical scale, such that the force exerted to the PVDF sensor equals the one to the scale.}
	\label{fig_08}
\end{figure}

Due to the different mechanical properties of injected cells, the exhibited injection forces also vary significantly, and are usually distributed in the range of $\mu$N-mN \cite{wei2018survey}. To measure the real-time injection forces accurately, the PVDF sensor is thus chosen as the micro force sensor with the model of LDTM-028K from TE Connectivity Ltd. Compared with other micro force sensors, like capacitive, optical, and piezoresistive types \cite{wei2018survey}, this kind of sensor is ideally suitable for cell injection force measurement thanks to the superior performance of low cost (about 5 dollars), wide range ($\mu$N-mN), high resolution ($\mu$N), and relatively simple structure, and has been successfully applied in lots of cell injection force sensing \cite{shen2007force,tan2008mechanical,xie2010force}. 
According to the calibration method in \cite{xie2009penetration}, the PVDF force sensor is calibrated with the sensitivity of 0.0102 mN/mV as shown in Fig. \ref{fig_08}, where a high-accuracy electrical scale with a sensing range of 500 g and a resolution of 0.001 g is used to provide the calibration data.

\begin{figure}[!t]
	\centering
	\includegraphics[width=3.2in]{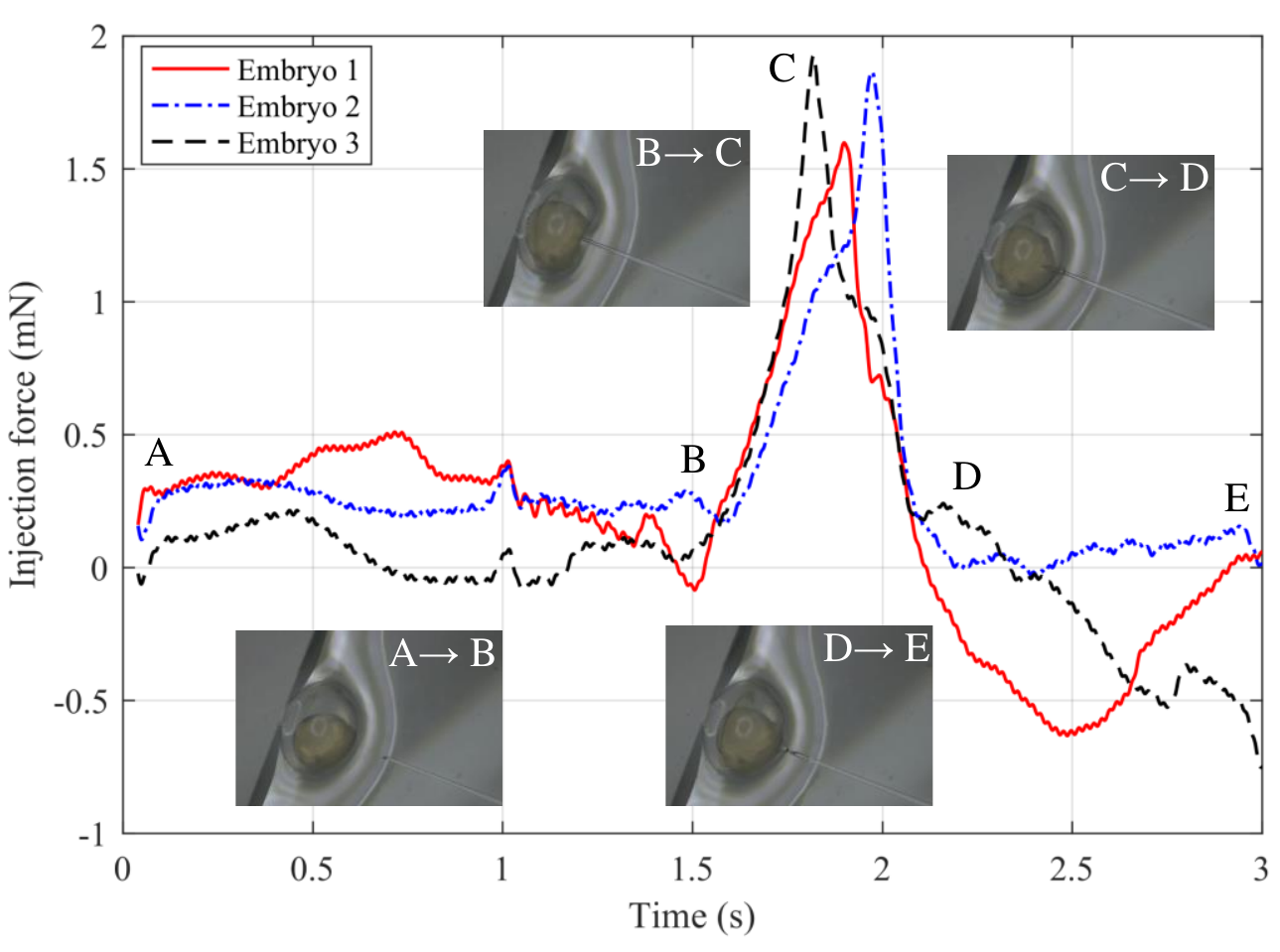}
	\caption{Measurement of injection force for different zebrafish embryos under a velocity $v=2.0$ mm/s and the corresponding main injection procedures: (A)-(B) the approaching phase; (B)-(C) the piercing phase; (C)-(D) the breakage phase of cell membrane; (D)-(E) the retracting phase of the needle.}
	\label{fig_09}
\end{figure}

As a class of widely used cell models in biomedical fields, zebrafish embryos are also introduced in this work due to their advantages of transparent and large embryos \cite{zhao2018review}. Normally,
the radii of zebrafish embryos are about 450-500 $\mu$m. Besides, for the different developmental stages of embryos, injection forces also differ from each other. So, to guarantee the consistency of the experimental objects, 50 zebrafish embryos at the pharyngula stage with radii of about 500 $\mu$m are collected as samples. Fig. \ref{fig_09} depicts some typical force response curves during microinjection and the corresponding main procedures, where the microscope views of the embryo at each phase are also shown. It is seen that for the three chosen embryos, the response curves are almost the same. Thus, in the following experiments, the injected embryos are considered to have similar mechanical properties without loss of generality. 

It is worth noting that the cell deformation is obtained by the feed movement of the end-effector from the initial time when the needle contacts the cell (e.g., Point B in Fig. \ref{fig_09}) to the final time when the needle punctures the cell (e.g., Point C in Fig. \ref{fig_09}). Specifically, the initial time is determined by the suddenly increasing point of the injection force response curve at the approaching phase, and the final time is by the peak point at the piercing phase. After determining the motion time and the injection speed, the feed movement is thus calculated.

\subsection{Parameter Identification}
\label{sec_5_2}

\begin{figure}[!t]
	\centering
	\includegraphics[width=3.5in]{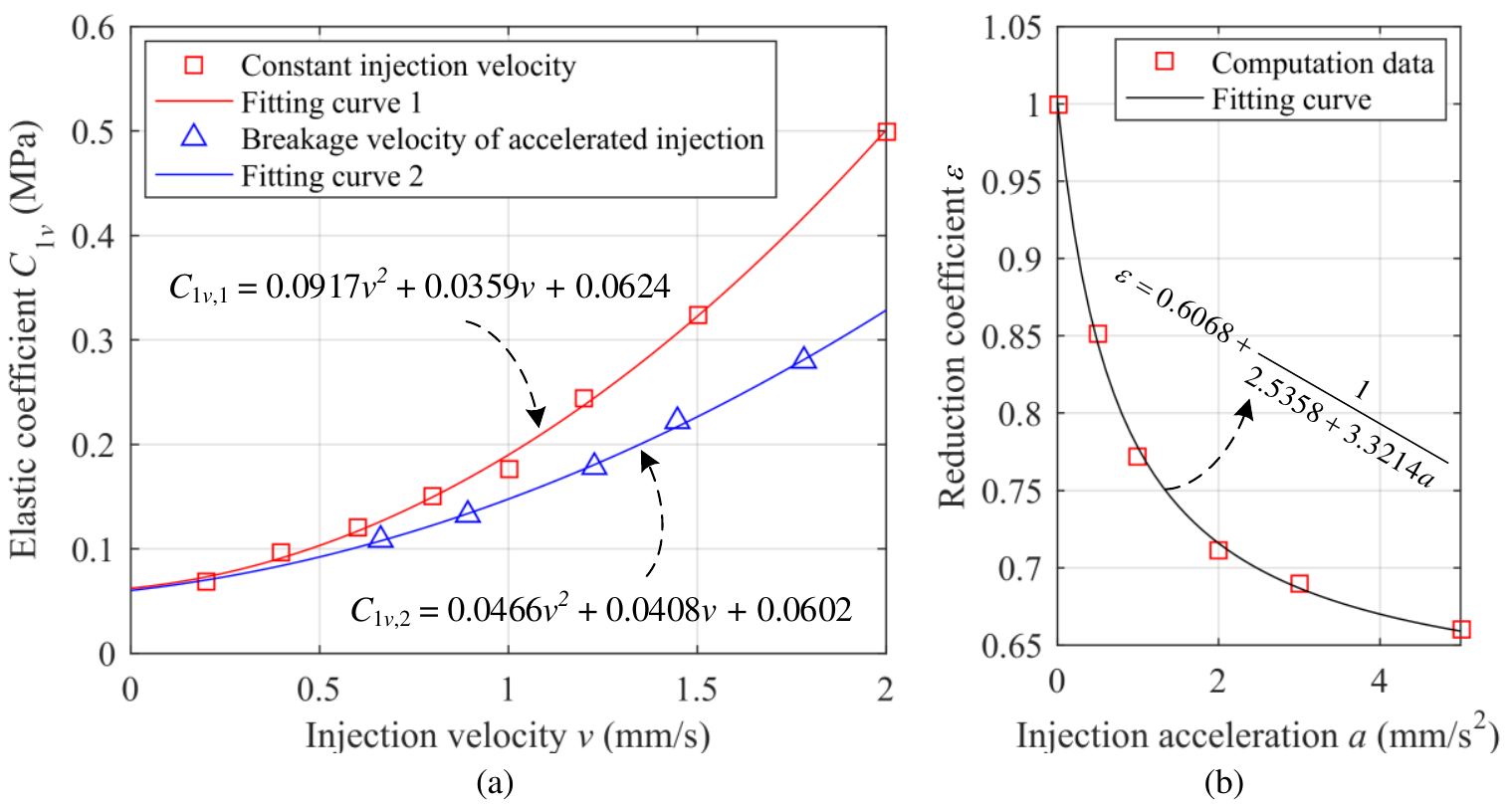}
	\caption{Parameter identification results of the elastic coefficient $C_{1v}$ (a) and the reduction coefficient $\varepsilon$ (b), where $C_{1v,1}$ represents the identified result under the constant injection velocities, and $C_{1v,2}$ represents the one under the accelerated injection.}
	\label{fig_010}
\end{figure}

In view of the proposed rate-dependent mechanical model, it needs to identify the relation between the injection speed and the elastic coefficient $C_1$ in Eq. (\ref{eq_027}). Here, a two-step identification method is selected, i.e., firstly, the constant injection velocity is applied to identify the parameters $\{ k_0, k_1, k_2 \}$, and then the accelerated injection is performed to identify the parameters $\{ g_0, g_1, g_2 \}$. To obtain the computation data of $C_1$ from the real mechanical model, the constant $\alpha$ for Mooney-Rivlin material in Eq. (\ref{eq_024}) is set to 0.2 according to \cite{tan2010characterizing}; the initial radius $r_0$ of embryos is observed by the microscope as about 500 $\mu$m; the thickness of cell membranes $h$ is obtained from \cite{tan2008mechanical} as $3$ $\mu$m; and the tip radius $\rho_0$ of the needle used in the experiments is selected as 50 $\mu$m. As for the speed data, through adjusting the input pulse frequencies of motors, different velocities and accelerations are provided to inject the cells. The identification results are presented in Fig. \ref{fig_010}. 

For the first step, different constant injection velocities are applied and the corresponding values of the elastic coefficient (denoted as $C_{1v,1}$) are calculated by the measured force and deformation data based on the proposed mechanical model. By the polynomial fitting, the parameters $\{ k_0, k_1, k_2 \}$ are identified as $k_0=0.0624$, $k_1=0.0359$, and $k_2=0.0917$. 

While for the second step, different injection accelerations are performed. By collecting the instantaneous breakage velocities of the membrane, injection force, and cell deformation, the elastic coefficient (denoted as $C_{1v,2}$) can be also calculated. It is found that $C_{1v,1}$ is reduced compared with $C_{1v,2}$ when the constant injection becomes the accelerated injection. This implies the influence of speed effect on the mechanical property of cells, and also gives a sufficient reason why the reduction coefficient $\varepsilon$ is added to correct $C_{1v,1}$.
Based on $C_{1v,1}$ and $C_{1v,2}$, the values of $\varepsilon$ at different injection accelerations are calculated from $\varepsilon=\frac{C_{1v,2}}{C_{1v,1}}$. Then, the parameters $\{ g_0, g_1, g_2 \}$ are accordingly identified as $g_0=0.6068, g_1=2.5358, g_2=3.3214$. After obtained the fitting curves of $C_{1v,1}$ and $\varepsilon$, the final elastic coefficient $C_1$ is identified by $C_1= \left( 0.0917 v^2 + 0.0359 v + 0.0624 \right) \left( 0.6068 + \frac{1}{2.5358 + 3.3214 a}\right)$.

\subsection{Experimental Results and Discussions}

\begin{figure}[!t]
	\centering
	\includegraphics[width=3.0in]{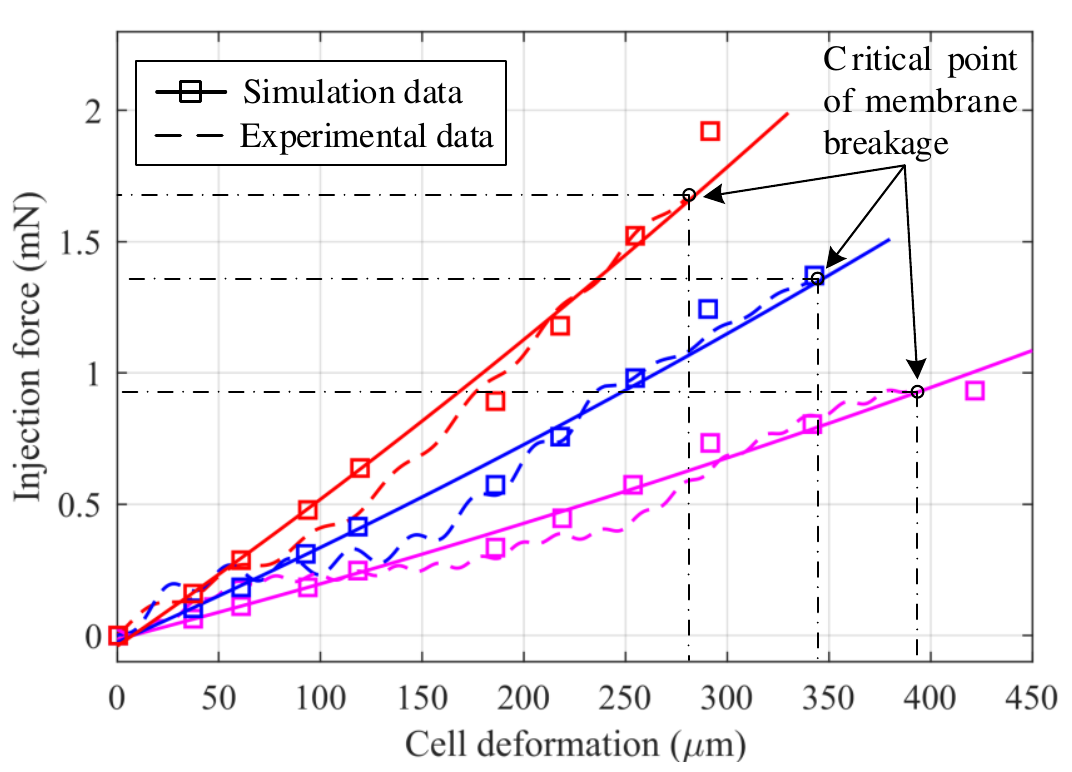}
	\caption{Comparisons between experimental results and simulation results using the proposed rate-dependent mechanical model at different injection velocities $v$. In this figure, the red color stands for the case of injection velocity $v=2$ mm/s, the blue color for $v=1.5$ mm/s, and the magenta color for $v=1.0$ mm/s.}
	\label{fig_011}
\end{figure}

According to the identification result of the elastic coefficient $C_1$, the relationship between the injection force and the cell deformation during microinjection can be obtained by using the proposed rate-dependent mechanical model. To validate the effectiveness, Fig. \ref{fig_011} shows comparisons between experimental results and simulation results at different injection velocities. It can be seen that the simulation results match the experimental results very well even though the injection velocities vary, which confirms that the proposed improved model is effective and essential to deal with the speed effect. 
 
\begin{figure}[!t]
	\centering
	\includegraphics[width=3.0in]{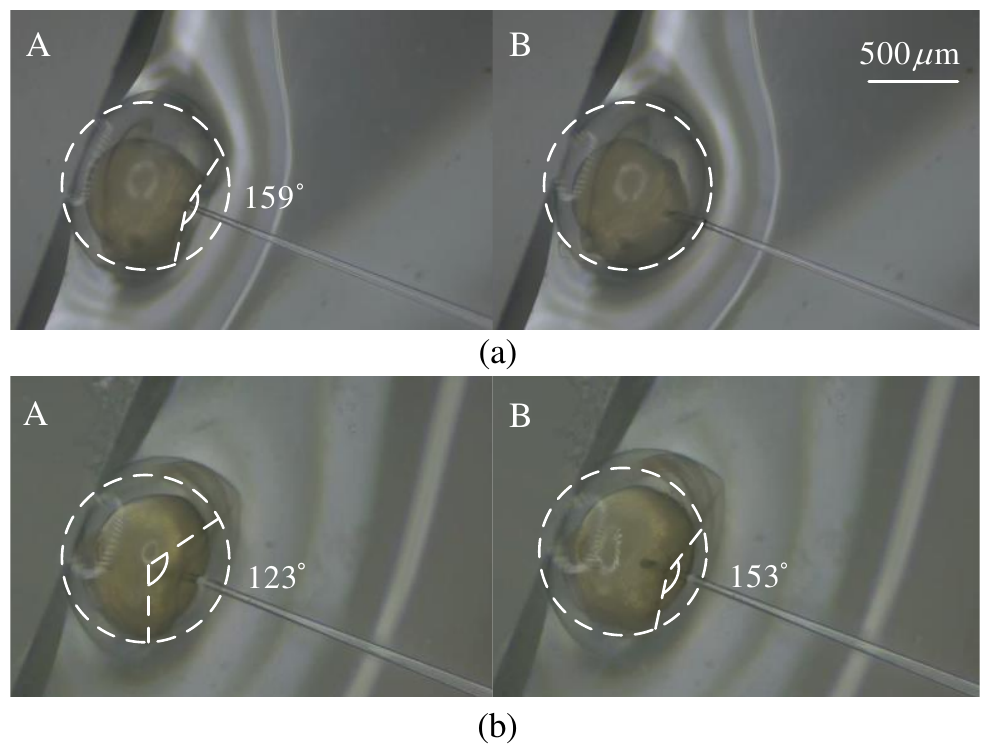}
	\caption{Cell deformations under the view of microscope at different injection speeds: (a) the fast constant velocity of 2 mm/s generated by the motion stage, (b) the slow non-uniform velocity generated by manual. In the figure, (A) and (B) show the cell deformations before and after breakage of the membrane, respectively. It shows that the robotic cell microinjection is better compared with the manual operation in terms of the deformation.}
	\label{fig_012}
\end{figure}

In addition, one can also observe from Fig. \ref{fig_011} that, with the increase of injection velocity, the total cell deformation decreases, but the injection force increases conversely. To further intuitively demonstrate the speed effect on the cell deformation, the microscope views for different injection speeds respectively generated by the motion stage and the manual operation are presented in Fig. \ref{fig_012}. It is obviously found that a faster velocity leads to a smaller deformation, and is more beneficial to reduce the degree of cell damage. However, it does not mean the faster the better, because the vibration of the end effector will be prominent under the high-speed injection. Thus, how to select the optimal speed trajectory will be a key problem to be solved in the future.

In view of the numerical simulation and the experimental results, one can safely conclude that the advantage of the proposed rate-dependent mechanical model is obvious over the traditional model \cite{tan2008mechanical} in terms of the speed effect. Since the mechanical model presented here can well predict the mechanical responses of the cells in the experiment, it exhibits great potential for biomedical applications, such as distinguishing abnormal cells from normal cells with high efficiency.

Nevertheless, it still exists some limitations in this work. Firstly, the designed injection device can not easily align with the center of the cell, and the relative position of the injection needle and the cell needs to be adjusted repeatedly, which is time-consuming and laborious. In addition, due to the high sensitivity of the PVDF sensor, the noise disturbance is significant in the experiments. In order to suppress the noise, the low-pass filter with the cut-off frequency of 20 Hz is used. But this correspondingly leads to a certain phase lag of the signal, so that the feed movement time can not be accurately obtained, and there exists some errors in the calculated cell deformation. Therefore, the microinjection system and micro force sensor will be further designed in the follow-up research to improve the usability and signal-to-noise ratio.

\section{Conclusion}
\label{sec_6}

This paper has proposed a new rate-dependent mechanical model based on the membrane theory for cell microinjection. The main contribution of this model is that the equilibrium equations between the injection force and the cell deformation are established with considering the speed effect for the first time. Thanks to the model, various mechanical responses can be predicted well during microinjection, including the force, the deformation, the distribution of tension and stress, and the deformed shape, etc. To verify the effectiveness of the model, a series of simulation analysis and experiments are also performed. The results demonstrate that the proposed model is in good agreement with the experimental data, even at different injection speeds. In the future, the microinjection device will be further improved, and the injection speed trajectory will be optimized.

\appendices
\section{Derivation of Eq. (\ref{eq_018})}
\label{app_1}
\setcounter{equation}{0}
\numberwithin{equation}{section}

According to Eq. (\ref{eq_011}), one can get
\begin{eqnarray}
\frac{\partial T_m }{\partial S} \rho + T_m \frac{\partial \rho }{\partial S} - T_c \frac{\partial \rho }{\partial S} + \sigma_m \rho  = 0   \nonumber   \\
\Rightarrow  \frac{\partial T_m }{\partial \psi} \frac{\partial \psi }{\partial \rho} \frac{\partial \rho }{\partial S} \rho + (T_m - T_c) \frac{\partial \rho }{\partial S} + \sigma_m \rho  = 0    \nonumber  \\
\Rightarrow  \frac{\partial T_m }{\partial \psi} \frac{1}{\rho'} \rho + (T_m - T_c) + \sigma_m \rho \frac{\partial S}{\partial \rho}  = 0    \nonumber   \\
\Rightarrow  \frac{\partial T_m }{\partial \psi} = \frac{\rho'}{\rho} (T_c - T_m) - \frac{\sigma_m \rho'}{\cos \theta}    \label{eq_a1}      
\end{eqnarray}

Since $T_m$ is the function of $\lambda_m, \lambda_c$ and $\psi$, the following equation holds
\begin{equation} \label{eq_a2}
\frac{\partial T_m }{\partial \psi} = \frac{\partial T_m }{\partial \lambda_m} \lambda_m' + \frac{\partial T_m }{\partial \lambda_c} \lambda_c'       
\end{equation}
Substituting Eq. (\ref{eq_a2}) into (\ref{eq_a1}), Eq. (\ref{eq_018}) is finally obtained.

\section*{Acknowledgment}
The authors would like to thank Mr. Liang Pan of Mettler Toledo (Changzhou in China) Measurement Technology Ltd. for the help in theoretical study and useful discussion.

\ifCLASSOPTIONcaptionsoff
  \newpage
\fi



\bibliographystyle{IEEEtran}

 
%\vspace{-10 mm}
\begin{IEEEbiography}[{\includegraphics[width=1in,height=1.25in,clip,keepaspectratio]{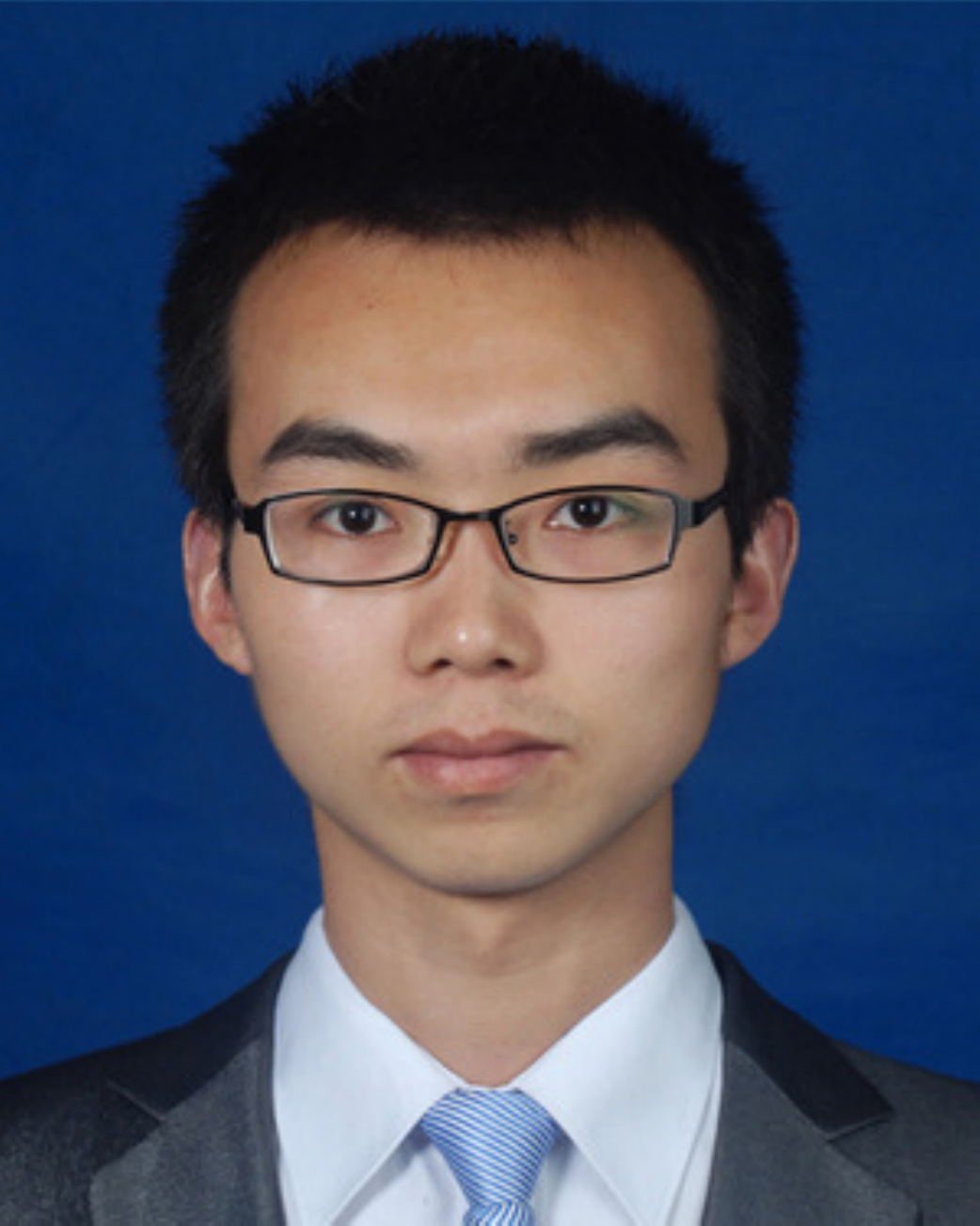}}]{Shengzheng Kang}
	received the B.S. degree in Mechanical Engineering and Automation in 2015, and the Ph.D. degree in Mechanical and Electrical Engineering in 2021, all from Nanjing University of Aeronautics and Astronautics, Nanjing, China.
	
	In 2018, he was a visiting Ph.D. student at the Department of Mechanical Engineering, Chiba University in Japan. From 2019 to 2021, he was a joint Ph.D. student sponsored by the China Scholarship Council (CSC) together with the Active Structures Laboratory in Department of Control Engineering and System Analysis, Universit\'{e} Libre de Bruxelles (ULB) in Belgium. Currently, Mr Kang is a Lecture at the School of Automation in Nanjing University of Information Science and Technology. His research interests include cell micromanipulation, compliant mechanisms and robot control.
\end{IEEEbiography}
\vspace{-10 mm}
\begin{IEEEbiography}[{\includegraphics[width=1in,height=1.25in,clip,keepaspectratio]{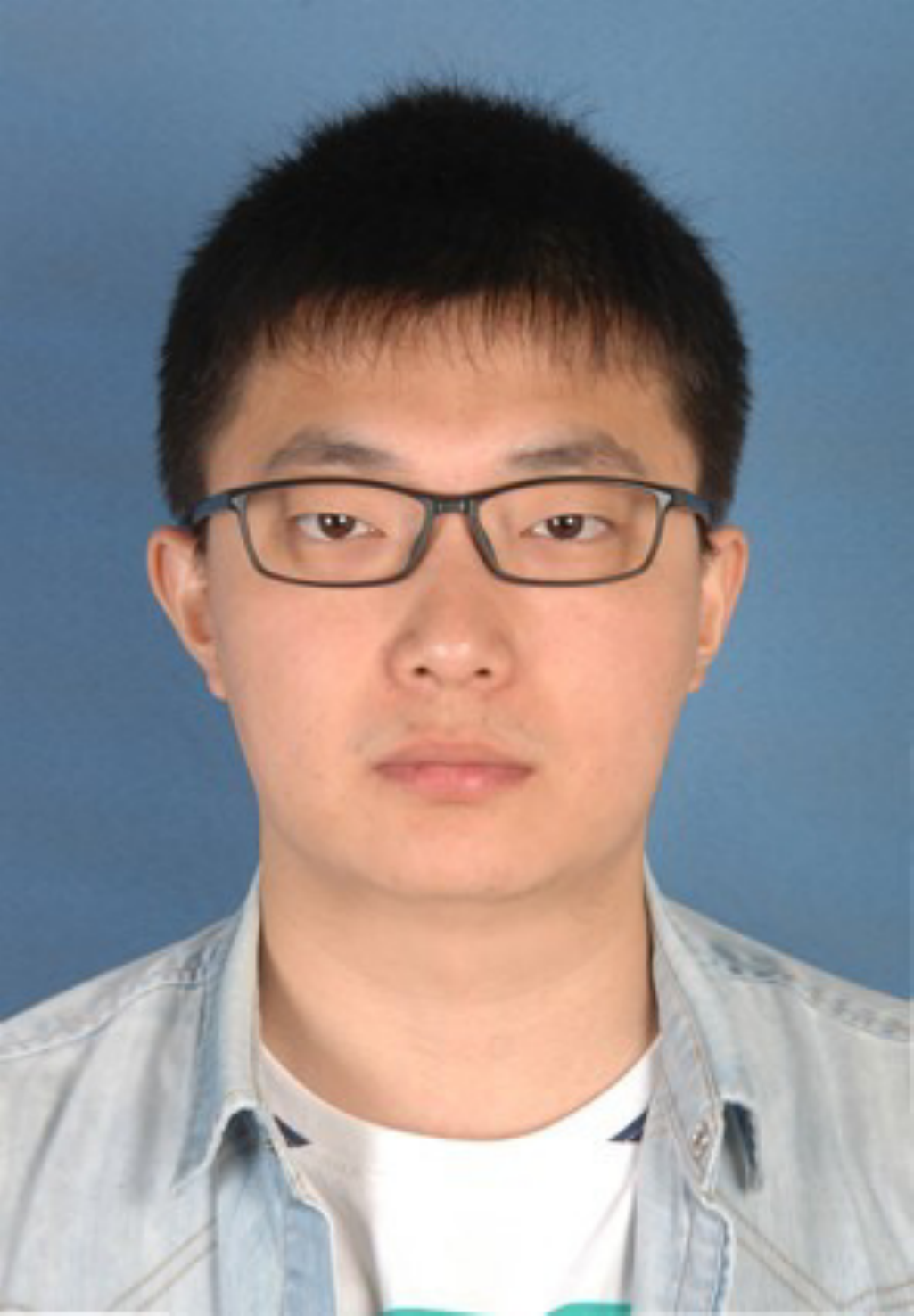}}]{Zhicheng Song}
	received the M.S. degree in Mechanical Engineering from Northeast Forestry University, Harbin, in 2020. He is currently pursuing a Ph.D. degree in Mechanical and Electronic Engineering with the Nanjing University of Aeronautics and Astronautics. 
	
	His main research interests include electromechanical system design of hand rehabilitation robots, bionic robots, and multi-body dynamics of parallel robots. 
\end{IEEEbiography}
\vspace{-10 mm}
\begin{IEEEbiography}[{\includegraphics[width=1in,height=1.25in,clip,keepaspectratio]{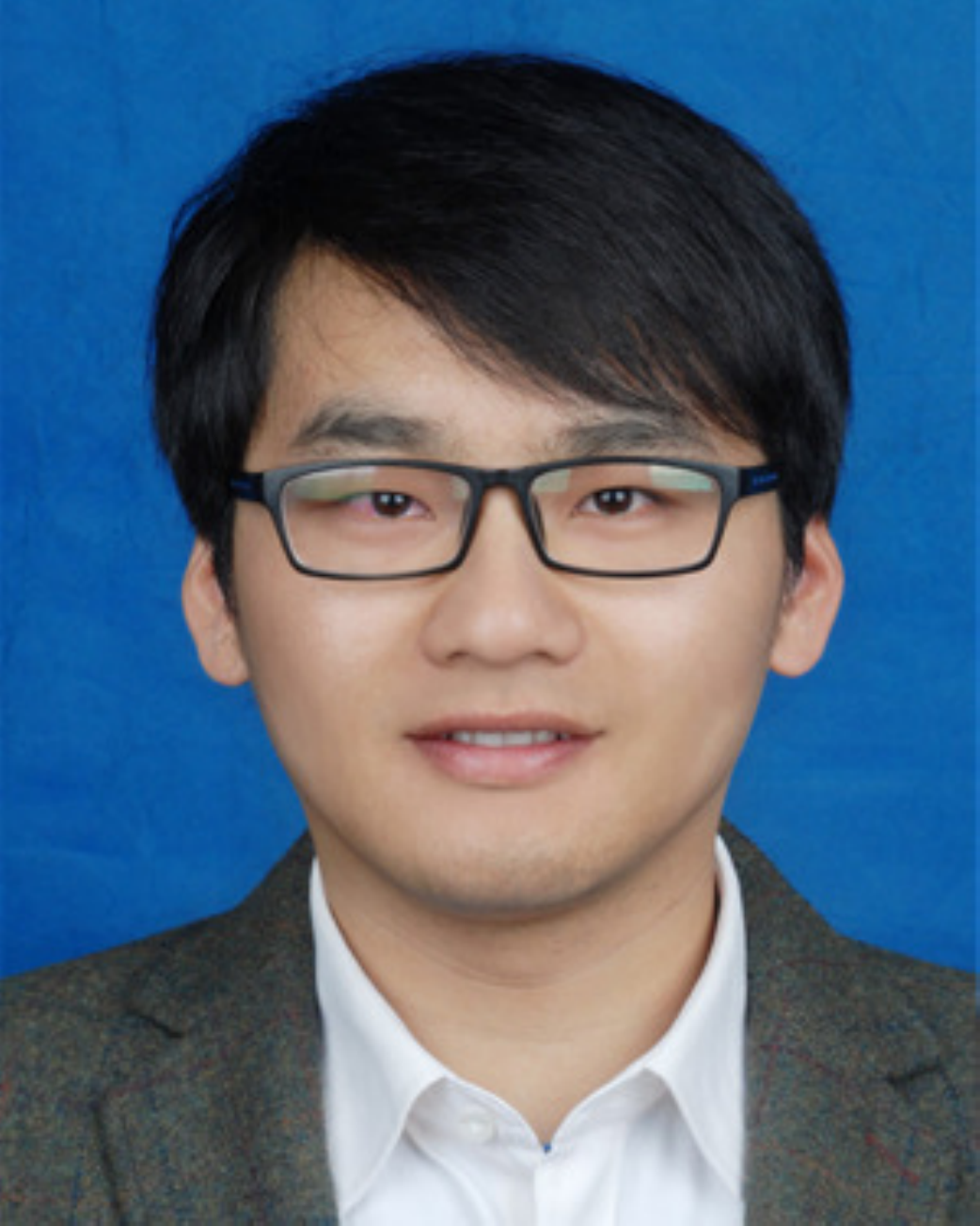}}]{Xiaolong Yang}
	received the B.S. degree in Mechanical Engineering and Automation in 2011 and the M.S. and Ph.D. degrees in Mechatronic Engineering in 2014 and 2018, all from Nanjing University of Aeronautics and Astronautics. 
	
	From 2018 to 2019, he was a Postdoctoral Fellow and Mechatronics Laboratory Director with the Lab of Biomechatronics and Intelligent Robotics at the Mechanical Engineering Department in the City University of New York, City College, US. Currently, Mr Yang is an Associate Professor at the School of Mechanical Engineering in Nanjing University of Science and Technology. His research interests include robotics, vibration control and exoskeletons.
\end{IEEEbiography}
\vspace{-10 mm}
\begin{IEEEbiography}[{\includegraphics[width=1in,height=1.25in,clip,keepaspectratio]{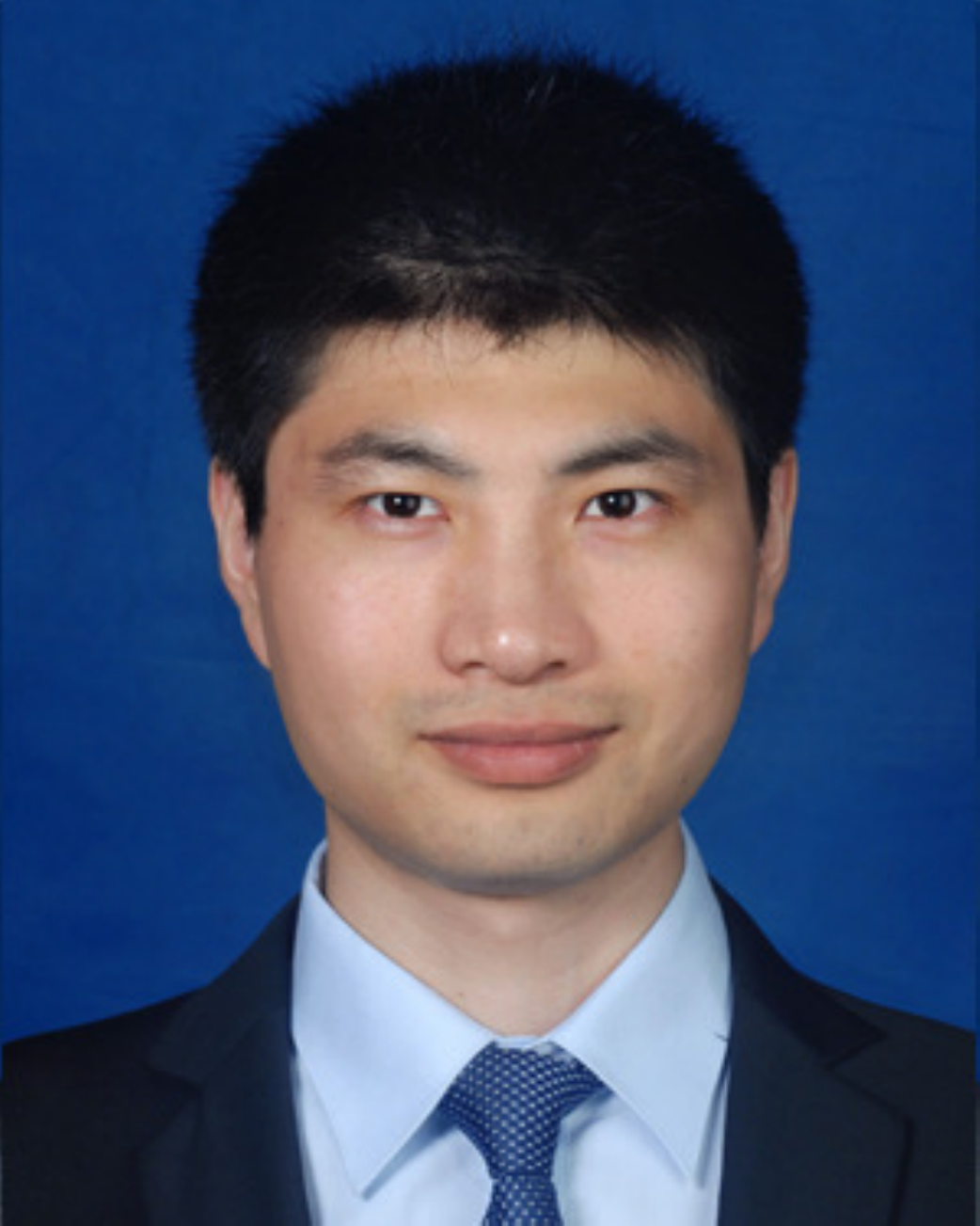}}]{Yao Li}
	received the B.S. degree in Mechanical and Electrical Engineering from Shandong Agricultural University, Taian, China, in 2012 and the M.S. and Ph.D. degrees in Mechanical and Electrical Engineering from Nanjing University of Aeronautics and Astronautics in 2015 and 2020.
	
	Mr Li is currently a Lecturer at the Industrial Center/School of Innovation and Entrepreneurship, Nanjing Institute of Technology from 2020. His research interests include parallel mechanisms and vibration control.
\end{IEEEbiography}
\vspace{-10 mm}
\begin{IEEEbiography}[{\includegraphics[width=1in,height=1.25in,clip,keepaspectratio]{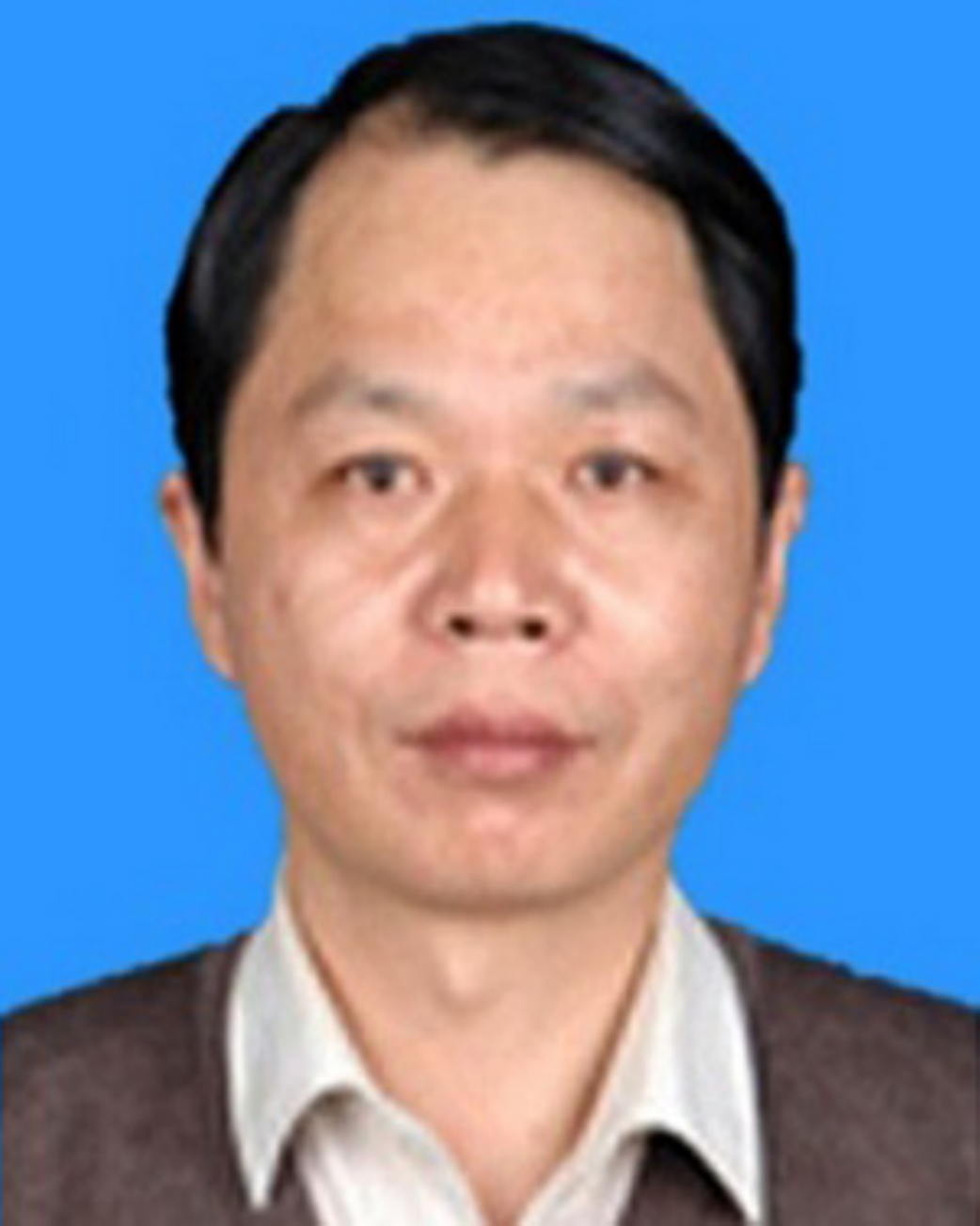}}]{Hongtao Wu}
	received the B.S. degree from Yanshan University, Hebei, China, in 1982, and the M.S. degree and Ph.D. degree from Tianjin University, Tianjin, China, in 1985 and 1992, respectively, all in Mechanical Engineering. 
	
	In 2000, he was a Research Associate with Hong Kong Polytechnic University. From 2001 to 2002, he was a Visiting Scholar at the Arizona State University. Since 1998, Mr Wu has been a Full Professor at the College of Mechanical and Electrical Engineering in Nanjing University of Aeronautics and Astronautics. His research interests include multibody dynamics, parallel mechanisms and robotics. 
\end{IEEEbiography}
\vspace{-10 mm}
\begin{IEEEbiography}[{\includegraphics[width=1in,height=1.25in,clip,keepaspectratio]{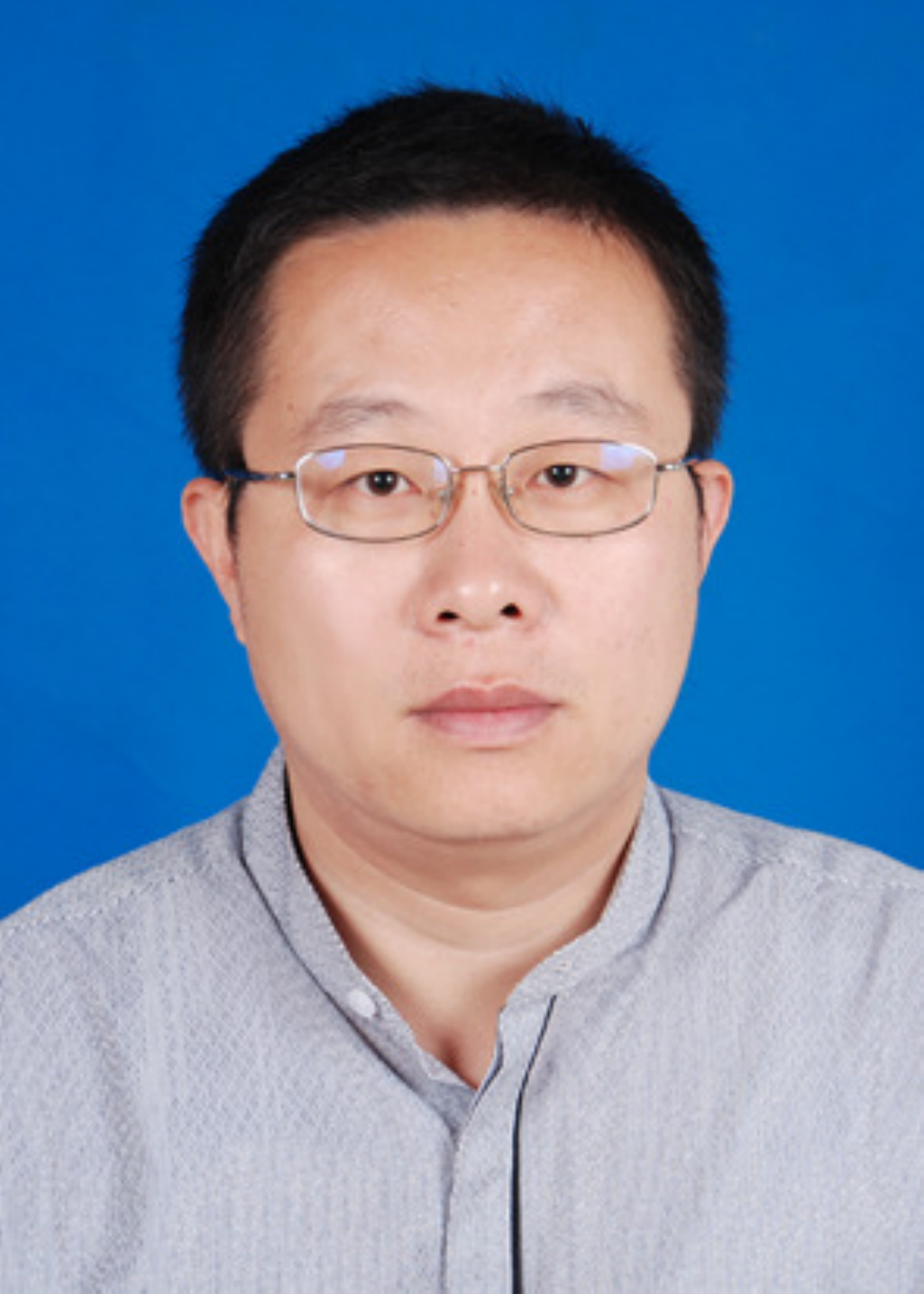}}]{Tao Li}
	received the Ph.D. degree from Southeast University, Nanjing, China. He is a Full Professor at the School of Automation in Nanjing University of Information Science and	Technology. His current research interests include fault detection and fault tolerant control for time-delay systems. He has published more than 70 papers in journals including Automatica, IEEE Transactions on Neural Networks, and IEEE Transactions On Systems, Man, And Cybernetics: Systems. He was a recipient of the Outstanding Young Scholar of Jiangsu Province, China. He completed the visiting scholar fellowships with the University of Alberta in Canada, the University of Western Sydney in Australia, the University of Hong Kong in China and the City University of Hong Kong in China.
\end{IEEEbiography}
%\vspace{-20 mm}

\end{document}